\documentclass[10pt,twocolumn,letterpaper]{article}

\usepackage{cvpr}              

\usepackage{graphicx}
\usepackage{amsmath}
\usepackage{amssymb}
\usepackage{amsfonts}
\usepackage{booktabs}
\usepackage{microtype}
\usepackage{algorithm}
\usepackage{algorithmic}
\usepackage{xcolor, colortbl}
\definecolor{mygray}{gray}{0.92}
\usepackage[pagebackref,breaklinks,colorlinks]{hyperref}

\usepackage{stfloats}

\renewcommand{\raggedright}{\leftskip=0pt \rightskip=0pt plus 0cm}
\usepackage{subcaption}
\usepackage{float}
\usepackage{bbm}
\usepackage{array}
\usepackage{pifont}
\usepackage{setspace}
\usepackage{makecell}
\usepackage{multirow}
\usepackage{multicol}

\usepackage{wasysym}

\newlength\savewidth

\usepackage{microtype}

\usepackage[capitalize]{cleveref}
\crefname{section}{Sec.}{Secs.}
\Crefname{section}{Section}{Sections}
\Crefname{table}{Table}{Tables}
\crefname{table}{Tab.}{Tabs.}

\def\cvprPaperID{1720} 
\def\confName{CVPR}
\def\confYear{2023}

\begin{document}

\title{SAP-DETR: Bridging the Gap between Salient Points and Queries-Based Transformer Detector for Fast Model Convergency}

\author{\large{Yang Liu$^{1,3}$ \thanks{This work was done when working as an intern at AI Lab, Lenovo Research, Beijing, China.}\quad
Yao Zhang$^{1,3}$ \footnotemark[1] \quad
Yixin Wang$^{2}$ \footnotemark[1] \quad
Yang Zhang$^{4}$\quad
Jiang Tian$^{4}$}\\
\large{Zhongchao Shi$^{4}$\quad
Jianping Fan$^{4}$\quad
Zhiqiang He$^{1,3,5}$ \thanks{Corresponding author.}}\\
$^{1}$Institute of Computing Technology (ICT), Chinese Academy of Sciences \quad
$^{2}$Stanford University \\
$^{3}$University of Chinese Academy of Sciences \quad
$^{4}$AI Lab, Lenovo Research \quad $^{5}$Lenovo Ltd. \\
{\tt\small \{liuyang20c,zhangyao215\}@mails.ucas.ac.cn \quad yxinwang@stanford.edu } \\
{\tt\small\{zhangyang20,tianjiang1,shizc2,jfan1,hezq\}@lenovo.com}
}

\maketitle

\begin{abstract}
Recently, the dominant DETR-based approaches apply central-concept spatial prior to accelerating Transformer detector convergency. These methods gradually refine the reference points to the center of target objects and imbue object queries with the updated central reference information for spatially conditional attention. However, centralizing reference points may severely deteriorate queries' saliency and confuse detectors due to the indiscriminative spatial prior. To bridge the gap between the reference points of salient queries and Transformer detectors, we propose \textbf{SA}lient \textbf{P}oint-based \textbf{DETR} (\textbf{SAP-DETR}) by treating object detection as a transformation from salient points to instance objects. Concretely, we explicitly initialize a query-specific reference point for each object query, gradually aggregate them into an instance object, and then predict the distance from each side of the bounding box to these points. By rapidly attending to query-specific reference regions and the conditional box edges, SAP-DETR can effectively bridge the gap between the salient point and the query-based Transformer detector with a significant convergency speed. Experimentally, SAP-DETR achieves 1.4$\times$ convergency speed with competitive performance and stably promotes the SoTA approaches by $\sim$1.0 AP. Based on ResNet-DC-101, SAP-DETR achieves 46.9 AP. The code will be released at \url{https://github.com/liuyang-ict/SAP-DETR.}
\end{abstract}

\section{Introduction}
\label{sec:010}

Object detection is a fundamental task in computer vision, whose target is to recognize and localize each object from input images. In the last decade, various detectors~\cite{ssd,fastercnn,fcos,ge2021yolox,retinanet,sparsercnn} based on Convolutional Neural Networks (CNNs), have received widespread attention and made significant progress. Recently, Carion \textit{et al.}~\cite{detr} proposed a new end-to-end paradigm for object detection based on the Transformer~\cite{attention}, called DEtection TRansformer (DETR), which treats object detection as a problem of set prediction. In DETR, a set of learnable positional encodings, namely object queries, are employed to aggregate instance features from the context image in Transformer Decoder. The predictions of queries are finally assigned to the ground truth via bipartite matching to achieve end-to-end detection. 

\begin{figure}[t]
    \centering
    \includegraphics[width=3.3in]{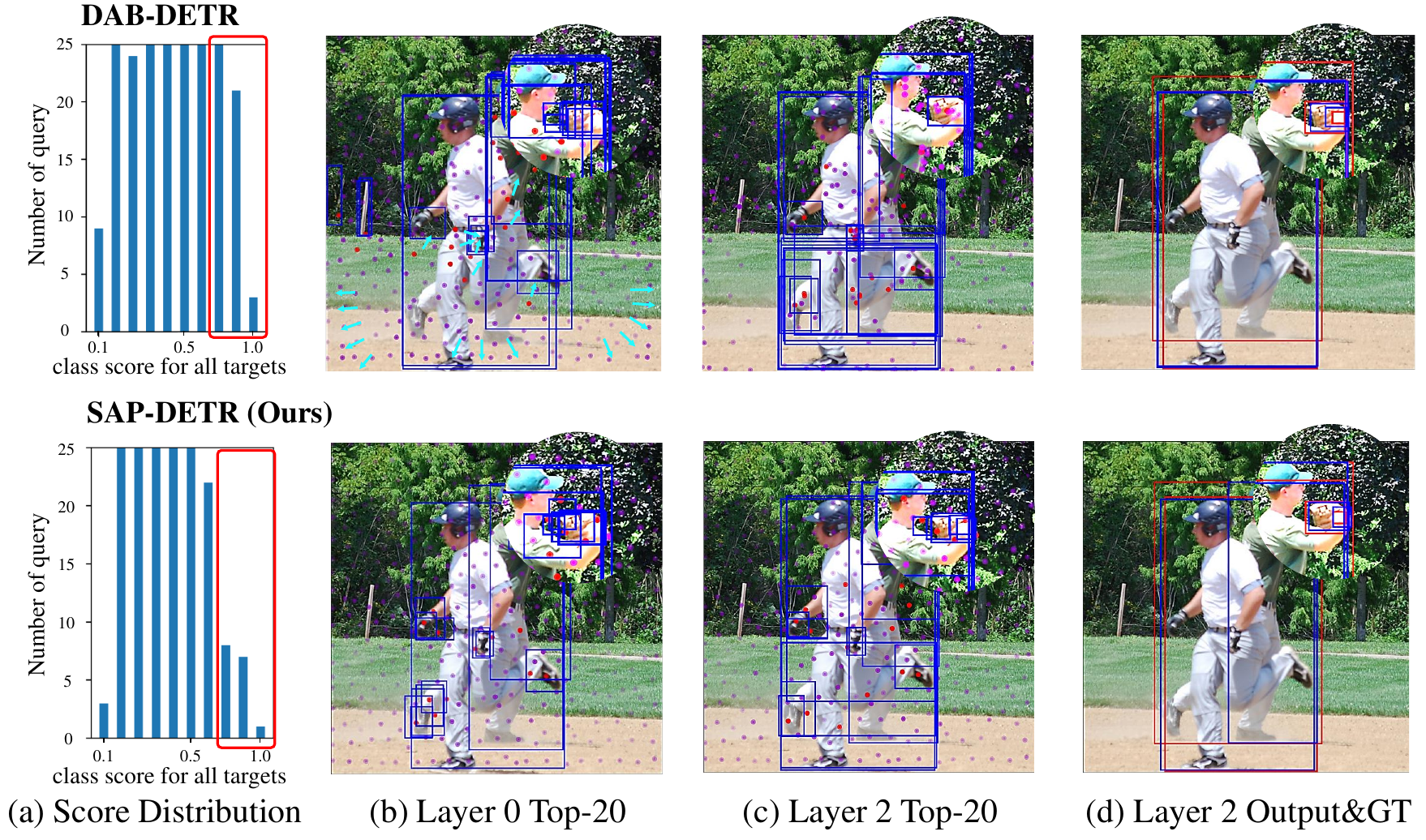}
	\caption{Comparison of SAP-DETR and DAB-DETR under 36 training epochs. (a) Statistics of the query count in different classification score intervals. (b) and (c) Distribution of reference points and the visualization of the query with top-20 classification score (blue proposal bounding boxes and red reference points) in different decoder layers. (d) Visualization of bounding boxes for positive queries (blue) and ground truth (red) during training process.}
	\label{fig:problem discription}
\end{figure}

Despite the promising results of DETR, its application is largely limited by considerably longer training time compared to conventional CNNs. To address this problem, many variants attempted to take a close look at query paradigm and introduced various spatial priors for model convergency and efficacy. According to the type of spatial prior, they can be categorized into implicit and explicit methods. The implicit ones~\cite{deformabledetr,smca,conditionaldetr} attempt to decouple a \textit{reference point} from the object query and make use of this spatial prior to attend to the image context features efficiently. The current state-of-the-arts (SoTAs) are dominated by the explicit ones~\cite{wang2021anchor,liu2022dab}, which suggest to instantiate a position with spatial prior for each query, i.e., explicit reference coordinates with a center point or an anchor box. These reference coordinates serve as helpful priors and enable the queries to focus on their expected regions easily. For instance, Anchor DETR~\cite{wang2021anchor} introduced an anchor concept (center point with different box size patterns) to formulate the query position and directly regressed the central offsets of the bounding boxes. DAB-DETR~\cite{liu2022dab} further stretched the center point to a 4D anchor box concept $[cx,cy,w,h]$ to refine proposal bonding boxes in a cascaded manner. However, instantiating the query location as a target center may severely degrade the classification accuracy and convergency speed. As illustrated in~\cref{fig:problem discription}, there exist many plausible queries~\cite{sun2021makes} with high-quality classification scores (\cref{fig:problem discription}(a) within red box) and box Intersection over Union (IoU, see the redundant blue boxes in~\cref{fig:problem discription}(b) and (c)), which only brings a slight improvement on precision rate but inevitably confuses the detector on the positive query assignments when training with bipartite matching strategy. This is because the plausible predictions are considered in negative classification loss, which severely decelerates the model convergency. As shown in~\cref{fig:problem discription}(b) and (c), the predefined reference point of the positive query may not be the nearest one to the center of the ground truth bounding box, and the reference points tend to be centralized or marginalized (cyan arrows in~\cref{fig:problem discription}(b)), hence losing the spatial specificity. With further insight into the one-to-one label assignment during the training process, we find that the query, whose reference point is closest to the center point, also has a high-quality IoU, but it still exists a disparity with the positive query in the classification confidence. Therefore, we argue that such a centralized spatial prior may cause degeneration of target consistency in both classification and localization tasks, which leads to inconsistent predictions.

Furthermore, the mentioned central point-based variants also have difficulties in detecting occluded objects, because their queries may be assigned to the ambiguous spatial prior with overlapping centers. For example, \cref{fig:problem discription}(d) shows that the baseman in front of the image is detected twice while the other is totally omitted when they are largely overlapped. One solution proposed in Anchor DETR~\cite{wang2021anchor} is to predefine different receptive fields (similar to the scaling anchor box in YOLO~\cite{yolov3}) for the position of each query. However, increasing the diversity of the receptive fields for each position query is unsuitable for non-overlapped targets, as it still generates massive indistinguishable predictions for one position as same as other center-based models.

To bridge these gaps, in this paper, we present a novel framework for Transformer detector, called SAlient Point-based DETR (SAP-DETR), which treats object detection as a transformation from salient points to instance objects. Instead of regressing the reference point to the target center, we define \textit{the reference point belonging to one positive query as a salient point}, keep this query-specific spatial prior with a scaling amplitude, and then gradually update them to an instance object by predicting the distance from each side of the bounding box. Specifically, we tile the mesh-grid referenced points and initialize their center/corner as the query-specific reference point. To disentangle the reference sparsity as well as stabilize the training process, a movable strategy with scaling amplitude is applied for reference point adjustment, which prompts queries to consider their reference grid as the salient region to perform image context attention. By localizing each side of the bounding box layer by layer, such query-specific spatial prior enables compensation for the over-smooth/inadequacy problem during center-based detection, thereby vastly promoting model convergency speed. Inspired by~\cite{smca,conditionaldetr,liu2022dab}, we also take advantage of both Gaussian spatial prior and conditional cross-attention mechanism, and then a salient point enhanced cross-attention mechanism is developed to distinguish the salient region and other conditional extreme regions from the context image features.

We bridge the gap between salient points and query-based Transformer detector by speedily attending to the query-specific region and other conditional regions. The extensive experiments have shown that SAP-DETR achieves superior convergency speed and performance. To the best of our knowledge, this is the first work to introduce the salient point based regression into end-to-end query-based Transformer detectors. Our contributions can be summarized as follows.

\textbf{\textit{1)}} We introduce the salient point concept into query-based Transformer detectors by assigning query-specific reference points to object queries. Unlike center-based methods, we restrict the reference location and \textit{define the point of the positive query as the salient one}, hence enlarging the discrepancy of query as well as reducing the redundant predictions (see \cref{fig:problem discription}). Thanks to the efficacy of the query-specific prior, our SAP-DETR accelerates the convergency speed greatly, achieving competitive performance with ~30\% fewer training epochs. The proposed movable strategy further boosts SAP-DETR to a new SoTA performance.  

\textbf{\textit{2)}} We devise a point-enhanced cross-attention mechanism to imbue query with spatial prior based on both reference point and box sides for final specific region attention.

\textbf{\textit{3)}} Evaluation over COCO dataset has demonstrated that SAP-DETR achieves superior convergency speed and detection accuracy. Under the same training settings, SAP-DETR outperforms the SoTA approaches with a large margin.

\section{Related Work}
\label{sec:020}
\noindent \textbf{Anchor-Free Object Detectors.}
Classical anchor-free object detectors can be grouped into center-based and keypoint-based approaches. The center-based approaches aim to localize the target objects based on the central locations~\cite{retinanet} or predefined ROI~\cite{fcos}. For example, FCOS~\cite{fcos} treated all points within the bounding box as positive ones to predict their distances from each side ($[\ell,t,r,b]$), and a centerness score was then considered to prohibit the low-quality prediction whose point is located near the border. Compared with FCOS, we also restrict the candidate queries within the bounding box but treat only one as positive to perform end-to-end object detection via an inner matching cost. 

The target of keypoint-based approaches is to localize the specific object locations and assign them to the predefined keypoints of the object for box localized training. For instance, diagonal corner points were considered in CornerNet~\cite{law2018cornernet}, center point was further grouped into CenterNet~\cite{centernet}, and ExtremeNet~\cite{zhou2019bottom} added some conjectural extreme points for object localization. These works showed an impressive performance, but the complicated keypoint matching may limit their upper bound. Our SAP-DETR takes the advantage of salient point regression to focus on the distinct regions without complicated point-based supervision.

\noindent \textbf{Query-Based Transformer Detectors.}
DETR~\cite{detr} pioneered a new paradigm of Transformer detector for end-to-end object detection without any post-processing~\cite{softnms}. In DETR, a new representation, namely object query, aggregates the instance features and then yields a detection result for each instance object~\cite{liu2021survey}. Following DETR, many votarists put efforts on the optimization of convergency and accuracy. 

Sun \textit{et al.}~\cite{sun2021rethinking} revealed that the main reason for slow convergency of DETR is attributed to the Transformer decoder, and they considered an encoder-only structure to alleviate such a problem. For in-depth understanding of the object query, one way is to generate a series of implicit spatial priors from queries to guide feature aggregation in cross-attention layers. SMCA~\cite{smca} applied a Gaussian-like attention map to augment the query-concerned features spatially. The \textit{reference point} concept was first introduced by Deformable DETR~\cite{deformabledetr}, where the sampling offsets are predicted by each reference point to perform deformable cross-attention. Following such a concept, Conditional DETR~\cite{conditionaldetr} reformulated the attention operation and rebuilt positional queries based on the reference points to facilitate extreme region discrimination. Another way is towards position-instantiation explicitly, where this position information enables to directly conduct positional query generation. Anchor DETR~\cite{wang2021anchor} utilized a predefined 2D anchor point $[cx,cy]$ to explicitly capitalize on the spatial prior during cross-attention and box regression. DAB-DETR~\cite{liu2022dab} extended such a 2D concept to a 4D anchor box $[cx,cy,w,h]$ and refined it layer-by-layer. 

The recent accelerating convergency methods are based on auxiliary queries for facilitating detector discrimination. DN-DETR~\cite{li2022dn} demonstrates the slow model convergency is mainly caused by the instability of bipartite matching, thus providing a denoising training to eliminate this issue. DINO~\cite{zhang2022dino} inherits this advance and further introduces negative queries to perform contrastive denoising. Group-DETR~\cite{chen2022group} proposes a group-wise one-to-many label assignment to match multiple positive object queries with more gradients for fast DETR convergency.

The most relevant approaches to ours are Point-DETR~\cite{chen2021points} and SAM-DETR~\cite{zhang2022accelerating,zhang2022semantic}. The former applied a point encoder for annotated point label infusion in teacher model, and the latter directly updated content embeddings by extracting salient points from image features for query-image semantic alignment. Unlike these concepts, we redefine the salient point from the perspective of the positive query's position and replace the center-concept prior with the query-specific position, thereby attending extreme regions, differentiating queries' saliency, and alleviating redundant predictions.

\section{Method}
\label{sec:030}

\begin{figure}[t]
    \centering
    \includegraphics[width=2.8in]{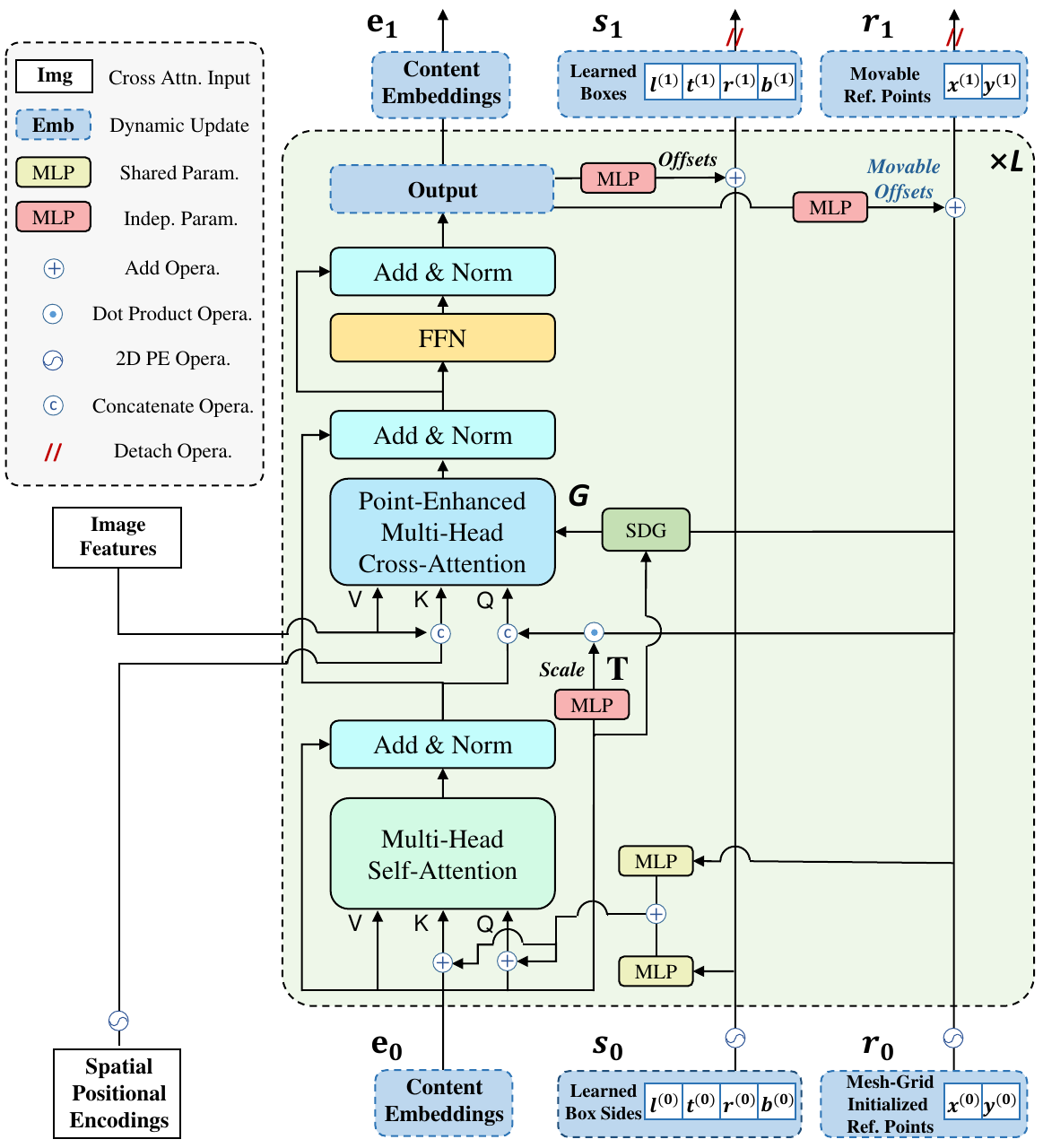}  
    \caption{Illustration of SAP-DETR. Each object query in SAP-DETR is assigned to a specific grid region and initialized by the corner/center of the grid as its reference point. A learnable 4D coordinate represents the distance from the four sides of the box to the reference point. Both reference points and box sides are served as positional encodings added/concatenated to content embeddings. All embeddings are refined to predict target objects gradually.}
    \label{fig:pipeline}
\end{figure}

We propose SAP-DETR to bridge the gap between salient points and query-based detectors. Following DETR, the extracted image features are fed into Transformer encoder after adding positional encodings, and then re-aggregated by object queries in Transformer decoder. In pursuit of the query-specific prior, we dispense a movable strategy for each query based on a fixed grid region. The query, whose reference region overlaps with ground truth objects, is allowed to predict the relative offsets from four sides of the bounding box to the points. Given the query-specific reference point and the proposal box sides, we propose salient point enhanced cross-attention mechanism to imbue query with spatial prior, thereby attending to extreme regions effectively. Additionally, we discuss two common issues in DETR-like models and address them for further improvements. 

\subsection{Salient Points-Based Object Detection}
\label{sec:031}
\noindent \textbf{Overview.} Previous methods~\cite{conditionaldetr, wang2021anchor, liu2022dab} normally decompose the object query into both content and position embeddings (queries), and form a center-based anchor point/box prior on the position ones. Unlike the central concept, we tile a fixed mesh-grid region, initialize their left-top corner as the reference points with 2D coordinate $\boldsymbol{r}\!=\!\{x, y\}\!\in\! [0,1]^2$, and instantiate a learnable 4D offset distance $\boldsymbol{s}\!=\!\{\ell, t, r, b\} \!\in\! [0,1]^4$ from the reference point to the sides of proposal bounding box for each object query. The object query can be referred as $\boldsymbol{q}\!=\!\{\mathbf{e};\boldsymbol{r}, \boldsymbol{s}\}$, where $\mathbf{e}\!\in\!\mathbb{R}^d$ is the content embedding with $d$ dimension. 
Instead of regressing the center, width, and height of a bounding box, we follow FCOS~\cite{fcos} and directly supervise the 4D offset from the four sides of a bounding box to the reference point. The final box prediction is formulated as $\boldsymbol{\hat{b}}\!=\! \{\hat{x}\!-\!\hat{\ell}, \hat{y}\!-\!\hat{t}, \hat{x}\!+\!\hat{r}, \hat{y}\!+\!\hat{b}\}$. Ideally, we here fix the reference point $\{\hat{x},\hat{y}\}=\{x,y\}$ (the movable update strategy is introduced in the next subsection) and only update the 4D box side prediction layer by layer. The prediction for each decoder layer can be calculated by
\begin{equation}\label{eq:01}
\begin{array}{c}
\Delta \boldsymbol{s}_l = \text{BoxHead}_l( \boldsymbol{s}_{l-1}, \mathbf{e}_{l-1}, \boldsymbol{r}_{l-1}),\vspace{1ex}\\ 
\hat{\boldsymbol{s}_l} = \sigma(\sigma^{-1}(\boldsymbol{s}_{l-1}) + \Delta \boldsymbol{s}_l), \quad
\boldsymbol{s}_l = \text{Detach}(\hat{\boldsymbol{s}_l}), \vspace{1ex}\\  
\boldsymbol{r}_l = \hat{\boldsymbol{r}_{l}} = \boldsymbol{r}_{l-1},\quad
\hat{\boldsymbol{b}_l} = \{\hat{\boldsymbol{r}_l}-\hat{\boldsymbol{s}_l}[:2],\hat{\boldsymbol{r}_l}+\hat{\boldsymbol{s}_l}[2:]\},
\end{array}
\end{equation}
where $\sigma$ and $\sigma^{-1}$ are the sigmoid and inverse sigmoid operation, respectively. $\Delta \boldsymbol{s}_l$ denotes the side offset prediction. $\hat{\boldsymbol{s}}_l$, $\hat{\boldsymbol{r}}_l$, and $\hat{\boldsymbol{b}}_l$ are the predicted side distance, reference points, and box location from the $l$ decoder layer, respectively. The BoxHead$_l$ is the prediction head following the layer-$l$ decoder, which is independent between different decoder layers in our settings. Detach operation follows DAB-DETR~\cite{liu2022dab}.

During the training process, each query is only allowed to predict the bounding boxes that overlap its reference region. We adapt this rule into the one-to-one bipartite matching process via an inner matching cost $\mathcal{L}_{\text{inner}}$. Given $N$ queries $\mathbf{Q}\!=\!\{ \boldsymbol{q}_1, \boldsymbol{q}_2, \cdots, \boldsymbol{q}_N\}$ and $M$ ground truth objects $\mathbf{G}\!=\!\{ \boldsymbol{g}_1, \boldsymbol{g}_2, \cdots, \boldsymbol{g}_M\}$, the $\mathcal{L}_{\text{inner}}(\boldsymbol{g}_i,\boldsymbol{q}_{j})$ of each query-box pair is a step function to penalize the reference point $\boldsymbol{r}_{j}$ of $\boldsymbol{q}_{j}$ with value of $k$ when $\boldsymbol{r}_{j}$ is outside the bounding box of $\boldsymbol{g}_i$. We denote $i\in[1,M]$ and $j\in[1,N]$ as the index of query and ground truth, respectively. $k$ can be viewed as a penalty cost, and default to $10^5$. The final permutation of the one-to-one label assignment is formulated as
\begin{equation}\label{eq:02}
\begin{aligned}
    &\quad \:{\mathcal{L}_{\text{inner}}(\boldsymbol{g}_i,\boldsymbol{q}_{j}):=\mathbf{k}_{\boldsymbol{r}_{j}\notin\boldsymbol{g}_i}},
    \\
    &\hat{\eta}\!=\!\mathop{\text{argmin}}\limits_{\eta \in \mathfrak{Y}_N }\sum\limits_{i}^{N}\mathcal{L}_{\text{match}} + \mathcal{L}_{\text{inner}} , 
\end{aligned}
\end{equation}
where $\mathcal{L}_{\text{match}}$ is the original pair-wise matching cost consist of both classification and localization costs~\cite{detr}. $\eta \!\in\!\mathfrak{Y}_N$ is a permutation of $N$ elements for bipartite matching.

\begin{figure}
\centering
\includegraphics[width=3.3in]{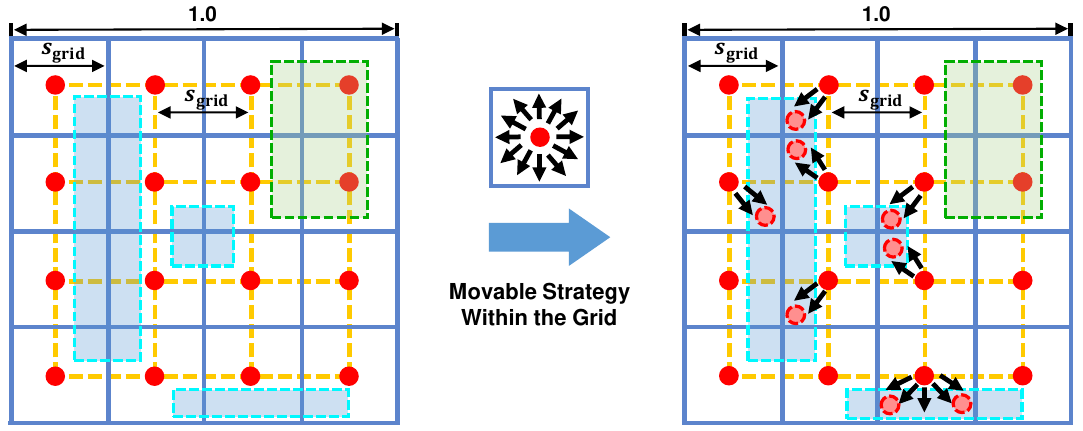}
\caption{Movable reference point. The reference points are initialized by the center/corner points of the mesh-grid. Based on the inner loss, only the green dashed box can be predicted by the inner points when reference points are fixed. By moving the reference points within their grid, the blue dashed boxes can be detected accurately without extended searching space.}
\label{fig:remoable}
\end{figure}

\noindent \textbf{Movable Reference Point.} Due to the sparseness of the fixed reference point, some small and slender objects may be indistinguishable when there is no reference point inside these objects. Despite the bipartite matching forcing each object to be assigned to one object query, the positive query, whose reference point is outside the assigned bounding box, is unable to accurately regress the distance from each side by a value between 0 and 1. One straightforward solution is to adjust the locations of reference points inside the ground truth bounding boxes to ensure that each object can be detected by an inner reference point. Similar to the aforementioned box refinement, we first perform a movable reference point design to dynamically update the reference points of each query layer by layer. However, such a full-image point regression inevitably expands the search space as vast variable determinations, causing the final reference point to be trapped in an unexpected corner of the bounding box. To reduce the training search spaces, we scale the offset amplitude of points within their specific grid regions, as illustrated in~\cref{fig:remoable}. Such an operation limits the range of offset values, and hence prevents a large searching space. It is implemented by applying the sigmoid activation $\sigma$ and multiplying a scale factor $\boldsymbol{s}_{\text{grid}}$ whose value equals to the height and width of one grid. The update process of the reference points is formulated as 
\begin{equation}\label{eq:03}
\begin{array}{c}
\Delta \boldsymbol{r}_l^\prime = \text{PointHead}_l( \boldsymbol{s}_{l-1}, \mathbf{e}_{l-1}, \boldsymbol{r}_{l-1}), \vspace{1ex} \\
\Delta \boldsymbol{r}_l = \sigma(\sigma^{-1}(\boldsymbol{r}_{l-1}-\boldsymbol{r}_{0}) + \Delta \boldsymbol{r}_l^\prime), \vspace{1ex}\\ 
\hat{\boldsymbol{r}_l} = \boldsymbol{r}_{0}+\Delta \boldsymbol{r}_l \cdot \boldsymbol{s}_{\text{grid}}, \quad
\boldsymbol{r}_l = \text{Detach}(\hat{\boldsymbol{r}_l}),
\end{array}
\end{equation}
where $\Delta \boldsymbol{r}_l^\prime$ and $\Delta \boldsymbol{r}_l$ are the predicted offsets from $\boldsymbol{r}_l$ to both $\boldsymbol{r}_{l-1}$ and $\boldsymbol{r}_0$ before the sigmoid activation $\sigma$, respectively. 

\subsection{Salient Point Enhanced Cross-Attention}
\label{sec:033}
In cross-attention layers, existing center-based methods are limited to the attention on both center and sides of the ground truth bounding box, causing detector confusion among the queries with the same center and side attention. To this end, we expect the queries to focus on their specific regions based on the reference points, four box sides, and other conditional regions in different heads. Accordingly, we consider an improved Gaussian~\cite{smca} $\textbf{G}$ and conditional attention~\cite{conditionaldetr} $\textbf{A}_{\text{peca}}$ to enhance query specificity and spatially extreme region discrimination. The final attention map $\textbf{A}_{\text{cross}}$ is the sum 
of the two attentions $\textbf{A}_{\text{cross}}=\textbf{G}+\textbf{A}_{\text{peca}}$.

\noindent \textbf{Side Directed Gaussian (SDG).} Similar to the movable strategy, we enforce the predicted Gaussian attention to be inside the proposal bounding box to reduce the searching space. Given a reference point $\boldsymbol{r}$, the offset scales $\mathbf{o} \in [-1, 1]^{2}$ for $H$ heads are produced by a simple MLP with a tanh activation, and then multiply to the two sides of the proposal bounding box for head-specific point offset generation, where the direction is guided by the sign of the offset scales. The head-specific points are generated by Algorithm~\ref{alg1}.
For each head, the Gaussian-like spatial weight map $G_i$ effecting on each pixel $(x,y)$ of context features is then formulated as
\begin{equation}\label{eq:05}
    \vspace{1ex}\\ 
    G_i(x,y)\!=\!\text{exp}\left(\!\!-\dfrac{(x - c_{w,i})^2}{v_{w,i}^2}\!-\!\dfrac{(y - c_{h,i})^2}{v_{h,i}^2}\!\!\right).
\end{equation}

\noindent \textbf{Point Enhanced Cross-Attention (PECA).}
\label{sec:0332}
As aforementioned, the semantic class for the query is closely related to its referenced location in our SAP-DETR. To enhance the correlation between queries and their references, we concatenate the locations to the content queries after the sinusoidal positional encoding (PE). Take a close look at the conditional attention~\cite{conditionaldetr}, we find that the linear positional embedding mostly focuses on one box side in each attention head. So we introduce a more straightforward attention mechanism, where the four side coordinates are concatenated and assigned to the corresponding head for side attention. The process of PECA is formulated as
\begin{equation}\label{eq:06}
    \begin{aligned}
    &&\hspace{-0.6cm}\textbf{A}_{\text{peca}}\!=\!&\;\mathbf{e}_q\mathbf{e}_k^\top  + \mathbf{T}\text{PE}(\boldsymbol{r}_q) \text{PE}(\boldsymbol{r}_k)\!^\top \\
    &&\ &\;+\mathbf{T}g(\text{PE}(\boldsymbol{r}_q\!-\!\{\ell,t\}, \boldsymbol{r}_q\!+\!\{r,b\})) \text{PE}(\boldsymbol{r}_k)\!^\top
    ,
    \end{aligned}
\end{equation}
where $g$ is a linear layer mapping PE(4D) into PE(2D) to keep channel dimension consistency.  $\mathbf{T}$ is a scaling matrix that follows Conditional DETR~\cite{conditionaldetr}, and more details of $\mathbf{T}$ are available in \cref{app:02}.  
\begin{algorithm}[t]
    \small
    \caption{Side Directed Gaussian}
    \label{alg1}
    \begin{algorithmic}[1]
    \renewcommand{\algorithmicrequire}{\textbf{Input:}}
    \renewcommand{\algorithmicensure}{\textbf{Output:}}
    \REQUIRE Content embedding $\mathbf{e}$, reference point $\boldsymbol{r}$ and box $\boldsymbol{s}$. \\
    \ENSURE Head-specific points $\mathbf{c}\!=\!\{(c_{w,i}, c_{h,i})|i\!\in\!H\}$ and head-specific attention $\mathbf{v}\!=\!\{(v_{w,i}, v_{h,i)}|i\!\in\!H\}$.\\
    \STATE Predict offset scale and attention scale based on content embedding,
    $\mathbf{o}\!=\text{tanh}(\text{MLP}(\mathbf{e}))$, $\mathbf{v}\!=\!\text{MLP}(\mathbf{e})$;\\
    
    \FOR{$h \leftarrow 1 \in H$}      
        \STATE Select the index of direction guided by the sign of offset scale, $\{a, b\}\!=\!\text{sgn}(\mathbf{o}_i)+\{1,2\}$, \quad $a,b \in \{0,1,2,3\}$; \\
        \STATE According to the index of direction, predict head-specific point, $\mathbf{c}_i\!=\!\mathbf{o}_i \cdot \boldsymbol{s}[a, b] \!+\!\boldsymbol{r}$;\\
    \ENDFOR
    \RETURN $\mathbf{c}_i, \mathbf{v}_i, \ \forall i=1,...,H$
    \end{algorithmic}
\end{algorithm}

\subsection{SAP-DETR with Denoising Strategy} 
\label{sec:034}
To further explore the capability of our proposed SAP-DETR, we develop SAP-DN-DETR and SAP-DINO-DETR by adding the denoising auxiliary loss~\cite{li2022dn,zhang2022dino} into the training process. In the denoised SAP-DETR, the main difference from both DN-DETR and DINO lies in the noise design. Instead of the center point, we perform the box jittering and randomly sample a point from the intersection region between the original bounding box and the jittering one as the reference point. As the denoising strategy only serves as an auxiliary training loss increasing the training cost, the variants of denoising models are test-free whose Params and GFLOPs are the same as SAP-DETR models. 

\section{Experimental Results}
\label{sec:040}
\subsection{Implementation Details}
\label{sec:041}
We conduct the experiments on the COCO 2017~\cite{mscoco} object detection dataset, containing about 118K training images and 5K validation images. All models are evaluated by the standard COCO evaluation metrics. We follow the vanilla DETR~\cite{detr} structure that consists of a CNN backbone, a stack of Transformer encoder-decoder layers, and two prediction heads for class label and bounding box prediction. We use ImageNet-pretrained ResNet~\cite{resnet} as our backbone, and report results based on the ResNet and its $\times$1/16-resolution extension ResNet-DC. Unlike DAB-DETR~\cite{liu2022dab} sharing box and label head for each layer, we share the class head except the first layer and use an independent box head for the box regression of each layer (for more details please refer to \cref{app:04}). As the mesh-grid initialization for reference points in SAP-DETR, we consider the number of queries $N$ as a perfect square for uniform distribution. Unless otherwise specified, we use $N\!=\!400$ queries in the experiments. Precisely, we also provide a comparison under $N\!=\!300$ in \cref{tab:50epoch}, the standard setting in DETR-like models.

We adopt two different Transformer structures for experiments where a 3-layer encoder-decoder stack is evaluated to demonstrate our lightweight model efficacy compared with the traditional CNN detectors, and a 6-layer encoder-decoder stack is aligned with previous DETR variants to investigate the performance of large model. Both are trained on two training schemes: the 12-epoch and 36-epoch schemes with a learning rate drop after 11 and 30 epochs, respectively. All models are trained on the Nvidia A100 GPUs with batch size of 16 and 8 for ResNet and ResNet-DC, respectively. For more training details, please refer to \cref{app:07}.

\begin{figure*}[t]
    \centering
    \includegraphics[width=5in]{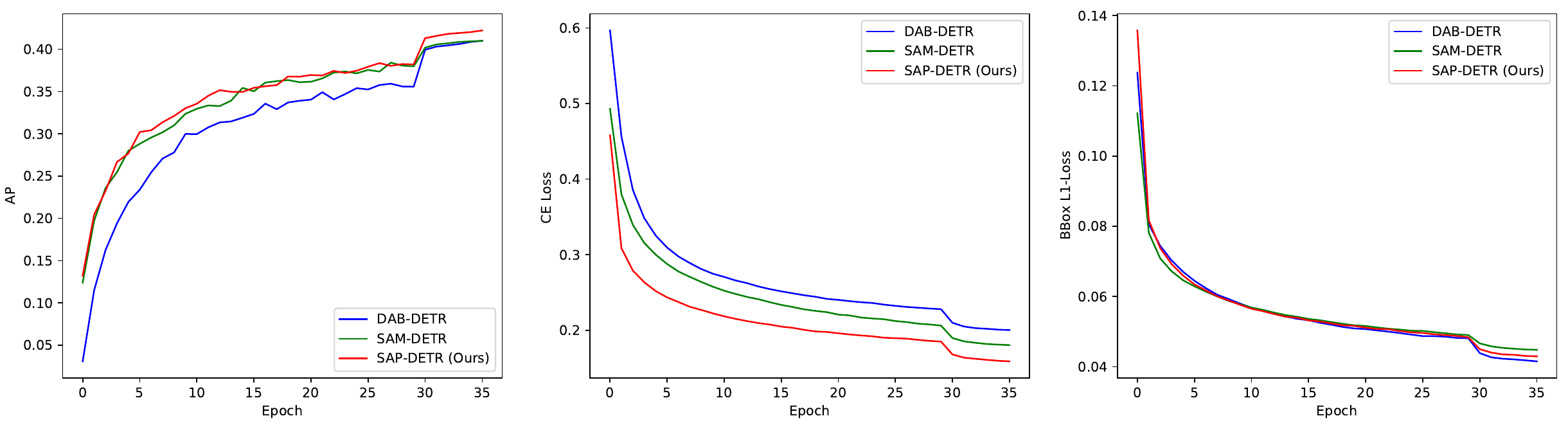}
    \vspace{-0.1cm}
    \caption{Comparison of performance and training losses curves.}
    \label{fig:36epoch-6layer-comparison}
\end{figure*}

\subsection{Main Results}
\label{sec:042}
As shown in~\cref{tab:31} and~\cref{tab:50epoch}, we comprehensively compare our proposed SAP-DETR with the traditional CNN detectors~\cite{fastercnn}, the original DETR~\cite{detr}, and other DETR-like detectors~\cite{deformabledetr,smca,conditionaldetr,wang2021anchor,liu2022dab,zhang2022accelerating} on COCO 2017 validation dataset. For in-depth analysis, we conduct the comparison in two aspects: model convergency and efficacy.

\noindent \textbf{Model Convergency.}
Compared with traditional CNN detectors, Transformer detectors are always subject to laborious training time. For example, under the same 12-epoch training scheme, Faster RCNN~\cite{fastercnn} still achieves good performance, but the mainstream DETR-like models may suffer from inadequate training and perform poorly without the help of auxiliary losses~\cite{li2022dn}. Under the 12-epoch training scheme, our proposed SAP-DETR can accelerate model convergency significantly, boosting DAB-DETR~\cite{liu2022dab} by 3.9 AP and 2.7 AP on 3-layer and 6-layer encoder-decoder structures, respectively. Compared with the current SoTA, our SAP-DETR also outperforms SAM-DETR~\cite{zhang2022accelerating} by $\sim$1.3 AP, with reducing $\sim$17\% parameters and $\sim$10\% GFLOPs. Take a close look at the training process, as illustrated in~\cref{fig:36epoch-6layer-comparison}, SAP-DETR conducts with rapid descent curves in both classification and box regression losses. Notably, there is a large margin in classification loss between ours and SoTA methods, which is benefited from the query-specific reference point, hence boosting model performance in early epochs. 

\noindent \textbf{Model Efficacy.}
To analyze model efficacy, we report results on long training epochs and high-resolution features in~\cref{tab:31}. Under the 36-epoch training scheme, SAP-DETR achieves superior performance among all single-scale Transformer detectors, \textit{especially on middle and large targets}. For example, SAP-DETR boosts DAB-DETR by 2.0 AP$_\text{M}$ and 4.1 AP$_\text{L}$ with 3-layer models, 1.0 AP$_\text{M}$ and 1.9 AP$_\text{L}$ with 6-layer models, which further verifies the effectiveness of our proposed salient point concept for overlapping object detection. Along with layer increase, a deficient upper-bound of SAM-DETR is exposed, with obviously lower 0.5 AP promotion compared to our 1.0 AP improvement. Persuasively, we also report the 50-epoch training results based on the 300-query setting. To align with our mesh-grid initialization strategy, we tile a $17\times 18$ mesh-grid (306 queries) for each reference point initialization. \cref{tab:50epoch} shows our main results and the most representative approaches with their original reported performance. Notably, SAP-DETR outperforms the current SoTAs with comparable costs based on all backbones. With low-resolution features ($\times1/32$), it significantly boosts both middle and large object detection accuracy.

\noindent \textbf{Combine with Other Fast Convergency Methods.}
As shown in Tab.~\ref{tab:sap-dn-detr}, we compare our SAP-DETR variants with the current fast
convergency methods~\cite{li2022dn,zhang2022dino,chen2022group}. With such a subtle modification, our SAP-DETR (grey rows) results in a significant performance improvement compared with the original methods (white rows). Under the 12-epoch training scheme, there exist 0.5-1.9 AP improvements on DN-DETR~\cite{li2022dn} and 0.7-1.6 AP improvements on Group-DETR~\cite{chen2022group}, but the promotions are slightly reduced when implemented on DINO~\cite{zhang2022dino}. We hypothesise that there exists the same effect between the negative query of contrastive denoising~\cite{zhang2022dino} and our query-salient reference point. Moreover, we observe that the performance improvements largely originate from the large object detection, especially based on ResNet-DC5 family backbones. We speculate that DETR may prefer the high-resolution features ($\times$1/16) rather than the low-resolution ones ($\times$1/32), and our SAP-DETR can distinguish the salient points accurately on the high-resolution, thereby taking full advantage of the large object detection.

\begin{table*}[ht]
\centering
  \setlength \tabcolsep{3pt}
    \begin{spacing}{0.79}
    \resizebox{0.95\linewidth}{!}{%
    \begin{tabular}{lccccccccc}
    \toprule
    Method & \#Epochs& \#Params(M) & GFLOPs& AP&AP$_{\text{50}}$&AP$_{\text{75}}$ & AP$_{\text{S}}$&AP$_{\text{M}}$&AP$_{\text{L}}$ \\
    \specialrule{0.9pt}{1.6pt}{3.6pt}
    \multicolumn{10}{l}{\textit{\textbf{3-Layer Encoder-Decoder Transformer Neck with ResNet-50 Backbone}}} \\
    \specialrule{0.5pt}{0.8pt}{1.7pt}
    DETR-R50~\cite{detr}            &  36  & 33   & 82     &   15.8    &  28.0     &   15.4  & 5.3 & 16.7 & 24.6  \\
    Deformable DETR-R50~\cite{deformabledetr} & 36 & 30   & 77    &    37.1   &  57.6     & 39.4   & 18.3 & 40.8 & 51.6  \\
    SMCA-DETR-R50~\cite{smca}       & 12 / 36  & -  & -  & 28.8 / 37.7 & 48.1 / 58.7 & 29.9 / 40.1 & 13.8 / 19.4 & 31.3 / 40.5 & 41.3 / 54.8  \\
    Conditional DETR-R50~\cite{conditionaldetr} & 12 / 36  & 40 & 82 & 29.6 / 37.1 & 48.7 / 57.9 & 30.7 / 39.0 & 13.0 / 17.6 & 32.3 / 40.3 & 43.1 / 55.0  \\
    Anchor DETR-R50~\cite{wang2021anchor}     & 12 / 36  & 31 & 79 & 30.8 / 37.6 & 51.1 / 58.7 & 31.8 / 39.7 & 14.3 / 18.8 & 34.1 / 41.5 & 44.3 / 53.5  \\
    DAB-DETR-R50~\cite{liu2022dab}        & 12 / 36  & 34 & 83 & 32.3 / 39.0 & 51.3 / 58.6 & 34.0 / 41.8 & 15.7 / 20.0 & 35.2 / 42.5 & 45.7 / 56.0  \\
    SAM-DETR-w/SMCA-R50~\cite{zhang2022accelerating} & 12 / 36  & 41 & 89 & 35.1 / 40.4   &   54.7 / 60.7    & 36.7 / 42.7 & 16.0 / 20.2 & 38.4 / 44.4 & 52.1 / 58.3     \\
    SAP-DETR-R50 (Ours) & 12 / 36  & 36 & 84 &  \textbf{36.2} / \textbf{41.2}   &   \textbf{56.2} / \textbf{61.6}    &  \textbf{37.9} / \textbf{43.4}  & \textbf{16.4} / \textbf{21.0}&\textbf{39.5} / \textbf{44.5} & \textbf{53.8} / \textbf{60.1}    \\
    \specialrule{0.9pt}{1.6pt}{3.6pt}
    \multicolumn{10}{l}{\textit{\textbf{6-Layer Encoder-Decoder Transformer Neck with ResNet-50 Backbone}}} \\
    \specialrule{0.5pt}{0.8pt}{1.7pt}
    DETR-R50~\cite{detr}              &  36 & 42 & 89    &    14.0   &    24.4   &  14.0  & 4.2 & 13.7 &  22.5   \\
    Deformable DETR-R50~\cite{deformabledetr}   &   36 & 34 & 81    &   38.0    &    58.2   & 40.4 & 18.5 & 41.7 &  54.2     \\
    SMCA-DETR-R50~\cite{smca}         &  12 / 36 &  - &  -  & 32.4 / 40.1 & 52.3 / 61.4 & 34.0 / 42.8 & 15.5 / 20.3 & 34.9 / 43.3 & 47.7 / 57.1  \\
    Conditional DETR-R50~\cite{conditionaldetr}  &  12 / 36 & 44 & 90  & 33.1 / 40.2 & 53.0 / 61.0 & 34.8 / 42.4 & 14.5 / 19.9 & 35.9 / 43.5 & 49.2 / 58.8 \\
    Anchor DETR-R50~\cite{wang2021anchor}       &  12 / 36 & 37 & 85  & 33.7 / 39.7 & 54.5 / 60.5 & 35.1 / 41.9 & 15.6 / 19.9 & 37.3 / 43.5 & 49.8 / 57.3 \\
    DAB-DETR-R50~\cite{liu2022dab}          &  12 / 36 & 44 & 92  & 34.9 / 41.0 & 55.5 / 61.7 & 36.4 / 43.4 & 16.2 / 21.3 & 38.4 / 44.7 & 51.5 / 58.9 \\
    SAM-DETR-w/SMCA-R50~\cite{zhang2022accelerating}   &  12 / 36 & 59 & 105 & 36.2 / 40.9 & 57.2 / 62.2 & 37.4 / 43.1 & 16.1 / 20.1 & 39.8 / 44.7 & 55.3 / 60.7    \\
    SAP-DETR-R50 (Ours)   &  12 / 36 & 47 & 94 & \textbf{37.5} / \textbf{42.2} & \textbf{58.5} / \textbf{62.7} & \textbf{39.2} / \textbf{44.6} & \textbf{17.3} / \textbf{22.6} & \textbf{40.6} / \textbf{45.7} & \textbf{55.4} / \textbf{60.8}  \\
    \bottomrule
    \end{tabular}}
    \end{spacing}
    \caption{Comparison between Transformer necks. Based on ResNet-50 backbone, all models are trained by the official source codes with their original settings and evaluated on COCO \texttt{val2017}. All models uses 400 queries except Anchor DETR, while Anchor DETR uses 200 queries with 2 pattern embeddings. GFLOPs and Params are measured by Detectron2\protect\footnotemark.}
    \label{tab:31}
\end{table*}%
\footnotetext{https:// github.com/facebookresearch/detectron2}

\begin{table*}[t]
\centering
  \setlength \tabcolsep{6pt}
    \begin{spacing}{0.7}
    \resizebox{0.95\linewidth}{!}{%
    \begin{tabular}{lcccccccccc}
    \toprule
    Method & \#Epochs& \#Params(M) & GFLOPs& AP&AP$_{\text{50}}$&AP$_{\text{75}}$ & AP$_{\text{S}}$&AP$_{\text{M}}$&AP$_{\text{L}}$&Infer. Time(s/img)$^\dag$\\
    \specialrule{0.9pt}{1.6pt}{3.6pt}
    \multicolumn{11}{l}{\textit{\textbf{ResNet-50 Backbone}}} \\
    \specialrule{0.5pt}{0.8pt}{1.7pt}
    Faster RCNN-FPN-R50~\cite{fastercnn,fpn}   & 108& 42 & 180 & 42.0 & 62.1 & 45.5 & 26.6 & 45.5 & 53.4&0.039 \\
    DETR-R50~\cite{detr}              & 500& 41 & 86  & 42.0 & 62.4 & 44.2 & 20.5 & 45.8 & 61.1&0.040 \\
    Deformable DETR-R50~\cite{deformabledetr}   & 50 & 34 & 78  & 39.4 & 59.6 & 42.3 & 20.6 & 43.0 & 55.5&0.043 \\
    SMCA-DETR-R50~\cite{smca}         & 50 & 42 & 86  & 41.0 & -    & -    & 21.9 & 44.3 & 59.1& 0.045\\
    Conditional DETR-R50~\cite{conditionaldetr}  & 50 & 44 & 90  & 40.9 & 61.8 & 43.3 & 20.8 & 44.6 & 59.2&0.057 \\
    Anchor DETR-R50~\cite{wang2021anchor}       & 50 & 39 & 85  & 42.1 & 63.1 & 44.9 & 22.3 & 46.2 & 60.0&0.050 \\
    DAB-DETR-R50~\cite{liu2022dab}          & 50 & 44 & \ \,\:90$^\dag$  & 42.2 & 63.1 & 44.7 & 21.5 & 45.7 & 60.3&0.059 \\
    SAM-DETR-w/SMCA-R50~\cite{zhang2022accelerating}   & 50 & 58 & 100 & 41.8 & 63.2 & 43.9 & 22.1 & 45.9 & 60.9&0.065 \\
    SAP-DETR-R50 (\textbf{Ours})   & 50 & 47 & 92  & \textbf{43.1} & \textbf{63.8} & \textbf{45.4} & \textbf{22.9} & \textbf{47.1} & \textbf{62.1}&0.063 \\
    \specialrule{0.9pt}{1.6pt}{3.6pt}
    \multicolumn{11}{l}{\textit{\textbf{ResNet-101 Backbone}}} \\
    \specialrule{0.5pt}{0.8pt}{1.7pt}
    Faster RCNN-FPN-R101~\cite{fastercnn,fpn}  & 108& 60 & 246  & 44.0 & 63.9 & 47.8 & 27.2 & 48.1 & 56.0&0.050 \\
    DETR-R101~\cite{detr}             & 500& 60 & 152  & 43.5 & 63.8 & 46.4 & 21.9 & 48.0 & 61.8&0.066 \\
    Conditional DETR-R101~\cite{conditionaldetr} & 50 & 63 & 156  & 42.8 & 63.7 & 46.0 & 21.7 & 46.6 & 60.9&0.070 \\
    Anchor DETR-R101~\cite{wang2021anchor}       & 50 & 58 & 150   & 43.5 & 64.3 & 46.6 & 23.2 & 47.7 & 61.4&0.068 \\
    DAB-DETR-R101~\cite{liu2022dab}         & 50 & 63 & \,\:157$^\dag$  & 43.5 & 63.9 & 46.6 & 23.6 & 47.3 & 61.5&0.072 \\
    SAP-DETR-R101 (\textbf{Ours})  & 50 & 67 & 158 & \textbf{44.4} & \textbf{64.9} & \textbf{47.1} & \textbf{24.1} & \textbf{48.7} & \textbf{63.1}&0.078 \\
    \specialrule{0.9pt}{1.6pt}{3.6pt}
    \multicolumn{11}{l}{\textit{\textbf{DC5-ResNet-50 Backbone}}} \\
    \specialrule{0.5pt}{0.8pt}{1.7pt}
    DETR-DC5-R50~\cite{detr}             & 500& 41 & 187  & 43.3 & 63.1 & 45.9 & 22.5 & 47.3 & 61.1&0.087 \\
    Conditional DETR-DC5-R50~\cite{conditionaldetr} & 50 & 44 & 195  & 43.8 & 64.4 & 46.7 & 24.0 & 47.6 & 60.7&0.093 \\
    Anchor DETR-DC5-R50~\cite{wang2021anchor}      & 50 & 39 & 151  & 44.2 & 64.7 & 47.5 & 24.7 & 48.2 & 60.6&0.069 \\
    DAB-DETR-DC5-R50~\cite{liu2022dab}         & 50 & 44 & \,\:194$^\dag$  & 44.5 & 65.1 & 47.7 & 25.3 & 48.2 & 62.3&0.094 \\
    SAM-DETR-w/SMCA-DC5-R50~\cite{zhang2022accelerating}  & 50 & 58 & 210 & 45.0 & 65.4 & 47.9 & 26.2 & 49.0 & \textbf{63.3}&0.126   \\
    SAP-DETR-DC5-R50 (\textbf{Ours})  & 50 & 47 & 197 & \textbf{46.0} & \textbf{65.5} & \textbf{48.9} & \textbf{26.4} & \textbf{50.2} & 62.6&0.116 \\
    \specialrule{0.9pt}{1.6pt}{3.6pt}
    \multicolumn{11}{l}{\textit{\textbf{DC5-ResNet-101 Backbone}}} \\
    \specialrule{0.5pt}{0.8pt}{1.7pt}
    DETR-DC5-R101~\cite{detr}             & 500& 60 & 253  & 44.9 & 64.7 & 47.7 & 23.7 & 49.5 & 62.3&0.101 \\
    Conditional DETR-DC5-R101~\cite{conditionaldetr} & 50 & 63 & 262  & 45.0 & 65.6 & 48.4 & 26.1 & 48.9 & 62.8&0.105 \\
    Anchor DETR-DC5-R101~\cite{wang2021anchor}      & 50 & 58 & 227    & 45.1 & 65.7 & 48.8 & 25.8 & 49.4 & 61.6&0.083 \\
    DAB-DETR-DC5-R101~\cite{liu2022dab}         & 50 & 63 & \,\:263$^\dag$  & 45.8 & 65.9 & 49.3 & 27.0 & 49.8 & 63.8&0.110 \\
    SAP-DETR-DC5-R101 (\textbf{Ours})  & 50 & 67 & 266 & \textbf{46.9} & \textbf{66.7} & \textbf{50.5} & \textbf{27.9} & \textbf{51.3} & \textbf{64.3}&0.130 \\
    \bottomrule
    \end{tabular}}
    \end{spacing}
    \caption{Comparison of Transformer necks with 300 queries on COCO \texttt{val2017}. All results are reported from their original paper. All models uses 300 queries except Anchor DETR, while Anchor DETR uses 100 queries with 3 pattern embeddings. All inference speeds are measured by a single Nvidia A100 GPU. $^\dag$ denotes the results are measured by ourselves.}
        \vspace{-0.15cm}
    \label{tab:50epoch}
\end{table*}%

\begin{table*}[ht]
\centering
  \setlength \tabcolsep{4.6pt}
    \begin{spacing}{0.9}
    \resizebox{0.95\linewidth}{!}{%
    \begin{tabular}{ccccccccccc}
    \toprule
    \multirow{2}[1]{*}{Backbone} & \multirow{2}[1]{*}{Epoch} & \multirow{2}[1]{*}{w/ SAP} & \multicolumn{2}{c}{DN-DETR~\cite{li2022dn}} &\quad& \multicolumn{2}{c}{DINO (Single-Scale)~\cite{zhang2022dino}} &\quad& \multicolumn{2}{c}{Group DETR~\cite{chen2022group}} \\
    \specialrule{0pt}{2pt}{0pt}
    \cline{4-5} \cline{7-8} \cline{10-11}
    \specialrule{0pt}{0pt}{2pt}
    &&&  AP\ /\ AP$_{\text{50}}$\ /\ AP$_{\text{75}}$ & AP$_{\text{S}}$\ /\ AP$_{\text{M}}$\ /\ AP$_{\text{L}}$ &\quad&
    AP\ /\ AP$_{\text{50}}$\ /\ AP$_{\text{75}}$ & AP$_{\text{S}}$\ /\ AP$_{\text{M}}$\ /\ AP$_{\text{L}}$  &\quad&
    AP\ /\ AP$_{\text{50}}$\ /\ AP$_{\text{75}}$ & AP$_{\text{S}}$\ /\ AP$_{\text{M}}$\ /\ AP$_{\text{L}}$ \\
    \specialrule{0.9pt}{1.6pt}{3.6pt}
    
    \multirow{2}[0]{*}{R50}    &\multirow{2}[0]{*}{12}&& 38.3  / 58.6   / 40.5   & 18.4  / 41.6 / 57.1 
    &\quad& 39.7  / 58.3   / 42.4   & 19.1  / 43.7 / 57.1 
    &\quad& 39.1  / \;\, - \;\, / \;\, - \;\,  & 19.7  / 42.5 / 56.8  \\
    
    &&\checkmark (\textbf{Ours})& \cellcolor{gray!25}\textbf{39.5}  / \textbf{59.7}   / \textbf{41.5}   &\cellcolor{gray!25} \textbf{18.7}  / \textbf{42.8} / \textbf{59.0} 
    &\cellcolor{gray!25}\quad &\cellcolor{gray!25}\textbf{40.0}  / \textbf{60.1}   / 42.1   &\cellcolor{gray!25} \textbf{20.2}  / 43.4 / \textbf{58.5} 
    &\cellcolor{gray!25}\quad &\cellcolor{gray!25}\textbf{39.8}  / \textbf{60.2}  / \textbf{42.0}   &\cellcolor{gray!25} \textbf{20.2}  / \textbf{43.5} / \textbf{58.6}  \\

    \specialrule{0.5pt}{0.8pt}{1.7pt}
    \multirow{2}[0]{*}{R101}   &\multirow{2}[0]{*}{12}&& 40.5  / 60.8   / 43.0   & 19.3  / 44.3 / 59.6 
    &\quad& 41.9  / 60.8   / 44.4   & 22.5  / 46.3 / 59.5 
    &\quad& \;\, - \;\,  / \;\, - \;\,   / \;\, - \;\,   & \;\, - \;\,  / \;\, - \;\, / \;\, - \;\, \\
    
    &&\checkmark(\textbf{Ours})& \cellcolor{gray!25}\textbf{41.0}  / \textbf{61.2}   / \textbf{43.4}   &\cellcolor{gray!25} \textbf{19.8}  / \textbf{45.3} / \textbf{60.0} 
    &\cellcolor{gray!25}\quad &\cellcolor{gray!25}41.5  / \textbf{61.4}   / 43.6   &\cellcolor{gray!25} 20.3 / 45.2 / \textbf{60.0}
    &\cellcolor{gray!25}\quad &\cellcolor{gray!25}\textbf{41.1}  / \textbf{61.5}   /\textbf{ 43.4}   &\cellcolor{gray!25} \textbf{20.5} / \textbf{45.5} / \textbf{59.4} \\
    
    \specialrule{0.5pt}{0.8pt}{1.7pt}
    \multirow{2}[0]{*}{R50-DC} &\multirow{2}[0]{*}{12}&& 41.7  / 61.4  / 44.1   & 21.2  / 45.0 / 60.2 
    &\quad& 43.6  / 61.4  / 47.0   & 24.8  / 47.3 / 59.5
    &\quad& 41.9  / \;\, - \;\, / \;\, - \;\,  & 23.3  / 45.6 / 58.4  \\
    
    &&\checkmark(\textbf{Ours})&  \cellcolor{gray!25}\textbf{43.6}  / \textbf{62.5}  / \textbf{46.2}   &\cellcolor{gray!25} \textbf{23.3}  / \textbf{47.3} / \textbf{61.0} 
    &\cellcolor{gray!25}\quad  &\cellcolor{gray!25}\textbf{44.0}  / \textbf{63.1}  / 46.5   &\cellcolor{gray!25} \textbf{24.8} / \textbf{47.3} / \textbf{61.1}
    &\cellcolor{gray!25}\quad  &\cellcolor{gray!25}\textbf{43.9}  / \textbf{63.2}  / \textbf{46.8}   &\cellcolor{gray!25} \textbf{24.5}  / \textbf{47.6} / \textbf{61.3}  \\
    
    \specialrule{0.5pt}{0.8pt}{1.7pt}
    \multirow{2}[0]{*}{R101-DC}&\multirow{2}[0]{*}{12}&& 43.4  / 61.9  / 47.2   & 24.8  / 46.8 / 59.4 
    &\quad&  45.4  / 63.5  / 49.2   & 26.4  / 49.5 / 61.1
    &\quad&  \;\, - \;\,  /\;\, - \;\,  / \;\, - \;\,   & \;\, - \;\,  / \;\, - \;\, / \;\, - \;\, \\
    
    &&\checkmark(\textbf{Ours})&\cellcolor{gray!25} \textbf{44.6}  / \textbf{63.9}  / \textbf{48.0}   &\cellcolor{gray!25} \textbf{25.5}  / \textbf{48.9} / \textbf{62.5} 
    &\cellcolor{gray!25}\quad  &\cellcolor{gray!25}\textbf{45.6}  / \textbf{64.5}  / 48.7   &\cellcolor{gray!25} 25.0  / \textbf{49.7} / \textbf{62.5}
    &\cellcolor{gray!25}\quad  &\cellcolor{gray!25}\textbf{44.4}  / \textbf{63.9}  /\textbf{47.4 }  &\cellcolor{gray!25}\textbf{25.9}  / \textbf{48.5} / \textbf{61.4} \\
    \bottomrule
    \end{tabular}}
    \end{spacing}
    \vspace{-0.1cm}
    \caption{Comparison with denoised methods on COCO dataset based on the 12-epoch training schedule and 300 object queries.}
    \vspace{0.3cm}
    \label{tab:sap-dn-detr}
\end{table*}%

\begin{table*}[t]
  \centering
  \setlength \tabcolsep{9pt}
  \begin{spacing}{0.7}
  \resizebox{0.95\linewidth}{!}{%
    \begin{tabular}{lcccccccccc}
    \toprule
    \multicolumn{1}{l}{Comment} &Movable & Inner Loss & PECA & SDG 
    & \multicolumn{1}{c}{AP} & \multicolumn{1}{c}{AP$_{\text{50}}$} & \multicolumn{1}{c}{AP$_{\text{75}}$} & \multicolumn{1}{c}{AP$_{\text{S}}$} & \multicolumn{1}{c}{AP$_{\text{M}}$} & \multicolumn{1}{c}{AP$_{\text{L}}$} \\
    \toprule
    SAP-DETR (\textbf{Ours})          & \checkmark & \checkmark & \checkmark & \checkmark  & \textbf{36.2} & 56.2     & \textbf{37.9} &   \textbf{16.4}    &   \textbf{39.5}    & \textbf{53.8} \\
    \quad $-$SDG             & \checkmark & \checkmark & \checkmark &             & 35.6  & 56.2  & 36.9  & 16.3  & 38.9  & 52.7 \\
    \quad $-$PECA            & \checkmark & \checkmark &            & \checkmark  & 34.8  & 55.5  & 36.0  & 15.7  & 37.3  & 52.0 \\
    \quad $-$PECA \& SDG     & \checkmark & \checkmark &            &             & 34.0  & 54.9  & 35.3  & 15.0  & 36.7  & 51.5 \\
    \quad $-$Movable       &            & \checkmark & \checkmark & \checkmark   & 35.2  & 55.4  & 36.8  & 15.8  & 38.5  & \textbf{53.8} \\
    \quad $-$Inner Loss      & \checkmark &            & \checkmark & \checkmark & 35.9  & \textbf{56.3}  & 37.4  & 16.2  & 39.3  & 52.5 \\
    \specialrule{0.5pt}{1.5pt}{2pt}
    DAB-DETR (Baseline)& \textbf{-} & \textbf{-} & \textbf{-} & \textbf{-}  & 32.3  & 51.3  & 34.0  & 15.7  & 35.2  & 45.7 \\
    \quad $+$Salient Point Concept & \textbf{-} & \textbf{-} & \textbf{-} & \textbf{-}  & 33.5  & 54.3  & 35.1  & 14.3  & 36.5  & 51.0 \\
    \bottomrule
    \end{tabular}}
    \end{spacing}
    \caption{Ablation on each components}
    \label{tab:effectiveness}%
\end{table*}%

\begin{table}[t]
  \centering
  \setlength \tabcolsep{2pt}
  \begin{spacing}{0.8}
    \centering
    \resizebox{1\linewidth}{!}{%
    \begin{tabular}{cccccc}
    \toprule
    Inner Cost ($\mathcal{L}_{\text{inner}}$)& Movable within Grid ($\boldsymbol{s}_{\text{grid}}$) & AP    & AP$_\text{S}$    & AP$_\text{M}$    & AP$_\text{L}$ \\
    \toprule
               &            &    35.9        & \textbf{17.0} &     38.8      &  52.7 \\
    \checkmark &            &    26.3        &   11.3        &     28.0      &  39.5 \\
    \checkmark & \checkmark &  \textbf{36.2} & 16.4          & \textbf{39.5} &  \textbf{53.8} \\
    \bottomrule
    \end{tabular}}
    \vspace{-0.1cm}
    \caption{Ablation on scaling factor of grid}
    \vspace{0.15cm}
    \label{tab:ablation_sfg}
    \end{spacing}
\end{table}%

\begin{table}[t]
  \centering
  \setlength \tabcolsep{7pt}
  \begin{spacing}{0.8}
    \centering
    \resizebox{1\linewidth}{!}{%
    \begin{tabular}{cccccc}
    \toprule
    \ \ PECA\  \   & Scaling Factor of SDG & AP    & AP$_\text{S}$    & AP$_\text{M}$    & AP$_\text{L}$ \\
    \toprule
               &            &  33.6          &  14.7         &     36.0      &  50.7 \\
    \checkmark &            &  35.7          &  \textbf{17.5}         &     38.8      &  52.6 \\
    \checkmark & \checkmark &  \textbf{36.2} & 16.4 & \textbf{39.5} &  \textbf{53.8} \\
    \bottomrule
    \end{tabular}}
    \caption{Ablation on scaling factor of SDG}
    \vspace{-0.1cm}
    \label{tab:ablation_sdg}
    \end{spacing}
\end{table}%

\subsection{Ablation Study}
\label{sec:050}
\noindent \textbf{Effectiveness of Each Component.}
To offer an intuitionistic comparison of model convergency for each component, \cref{tab:effectiveness} reports the effectiveness of them based on the 3-layer encoder-decoder structure and 12-epoch training scheme. \textbf{\textit{1).}} The proposed salient point concept based on content embeddings improves the performance from 32.3 AP to 33.5 AP compared to baseline DAB-DETR (row 8-9). Such a query-specific spatial prior enables queries to attend to their expected region from content features (see Figures in \cref{app:08}) and reduces the false detection rate on occluded and partial objects (see~\cref{fig:problem discription}(d)), hence boosting detection performance on middle and large objects. However, there exists a drop in small object detection (15.7 AP$_{\text{S}}$ \textit{vs.} 14.3 AP$_{\text{S}}$), for which we consider that the failure is mainly caused by the query sparsity. \textbf{\textit{2).}} Therefore, the movable strategy is applied to alleviate the constraint, improving the final model by 1.0 AP and 0.6 AP$_{\text{S}}$ (row 1 and 6). \textbf{\textit{3).}} Compared with the final model, the inner loss greatly improves the performance on high-quality AP$_{\text{75}}$ and large objects AP$_{\text{L}}$ detection (row 1 and 7), with just a slight drop on low-quality object AP$_{\text{50}}$. That is reasonable because the outside reference points are unable to localize objects accurately, and this phenomenon always exists in large objects. \textbf{\textit{4).}} For salient point enhanced cross-attention, both SDG and PECA serve as the essential components, independently emerging 0.8 AP and 1.6 AP improvements compared to the standard model (rows 2-4). Interestingly, there exists an effectiveness overlap on small objects, with only 0.1 AP$_{\text{S}}$ improvement when adding SDG to the equipped PECA model (row 1 and 2). We argue that the Gaussian-like map of SDG might be easily overlapped with PECA on small objects.

\noindent \textbf{Scaling Factor of Movable Strategy.}
We perform an ablation study on the scaling factor of the movable strategy and further investigate the effectiveness of the inner cost in~\cref{tab:ablation_sfg}. Notably, it is observed that there exists a conflict between the inner loss and the global search strategy, behaving a sharp drop when only reserving the inner loss. Furthermore, searching within the grid enables the detector to more attend to small objects and avoid a drastic deterioration in normal object detection. See \cref{app:05} for more detailed analyses.

\noindent \textbf{Scaling Factor of SDG.}
We also compare our side-directed manner with the standard offset prediction method in~\cref{tab:ablation_sdg}. Based on PECA, the side-directed scaling factor may limit the detector on small object detection but significantly promote the performance on other objects. This phenomenon would be broken without the help of PECA in which a precipitous decline is emerged on all-scale object detection (row 5 in~\cref{tab:effectiveness} \textit{vs.} row 1 in~\cref{tab:ablation_sdg}). We hypothesise that it because the predicted reference points may be outside of the proposal boxes, or even the region of the image.


\section{Conclusion}
\label{sec:060}
In this paper, we propose SAP-DETR for promoting model convergency by treating object detection as a transformation from the salient points to the instance objects. Our SAP-DETR explicitly initializes a query-specific reference point for each object query, gradually aggregates them into an instance object, and predicts the distance from each side of the bounding box. By speedily attending to the query-specific region and other extreme regions from contextual image features, it thus can effectively bridge the gap between the salient points and the query-based Transformer detector. Our extensive experiments have demonstrated that SAP-DETR achieves superior model convergency speed. With the same training settings, our proposed SAP-DETR outperforms SoTA approaches with large margins.

\section{Future Work}
\label{sec:070}
This point-based design for DETR-like models also comes with remaining issues, in particular regarding training with deformable attention, multi-scale features, and negative query design. \!\!\!Following current center-based methods working for similar issues, we expect future work to succe-ssfully address them for point-based design of \!SAP-DETR.\!\!\!\!\!.
\vspace{-0.9cm}
{\small
\bibliographystyle{ieee_fullname}
\bibliography{egbib}
}

\newpage
\appendix
\section*{Appendix}
\section{Comparison of DETR Family}
\label{app:01}

\begin{table*}[t]
\centering
  \setlength \tabcolsep{4.5pt}
    \begin{spacing}{1.3}
    \resizebox{0.95\linewidth}{!}{%
    \begin{tabular}{lcccccc}
    \toprule
    Method & Spatial Prior & Reference Coordinate & Target Prediction & Cross-Attn. & Reference Prior Update & Discriminative PE \\
    \specialrule{0.9pt}{1.6pt}{3.6pt}
    DETR             & No       & No    & [$cx,cy,w,h$] & Standard & & $\checkmark$      \\
    Deformable DETR  & Implicit & 4D    & [$dcx,dcy,w,h$] & Deformable Points & & $\checkmark$\\
    SMCA-DETR        & Implicit & 4D    & [$\Delta cx,\Delta cy,w,h$] & Gaussian Points & &   \\
    Conditional DETR & Implicit & 2D    & [$\Delta cx,\Delta cy,w,h$] & Conditional & &  \\
    Anchor DETR      & Explicit & 2D    & [$\Delta cx,\Delta cy,w,h$] &  Standard & &$\checkmark$   \\
    DAB-DETR         & Explicit & 4D    & [$\Delta cx,\Delta cy,\Delta w, \Delta h$] & Conditional &$\checkmark$&  \\
    SAM-DETR-w/SMCA  & Explicit & 4D    & [$\Delta cx,\Delta cy,\Delta w, \Delta h$] & Gaussian Points &$\checkmark$ & \\
    SAP-DETR (Ours)  & Explicit & 2D+4D & [$\Delta x,\Delta y,\Delta \ell, \Delta t,\Delta r, \Delta b $]& Conditional Side & $\checkmark$ & $\checkmark$ \\
    \bottomrule
    \end{tabular}}
    \end{spacing}
    \caption{Comparison of DETR-like models and our proposed SAP-DETR.}
    \label{tab:app01}
\end{table*}%
\cref{tab:app01} detailedly compares various representative properties for the DETR family. DETR~\cite{detr} follows the vanilla Transformer structure and leverages the learnable positional encodings to help Transformer distinguish paralleled input queries. However, such learnable positional encodings without any spatial prior help severely affect the convergency speed of the Transformer detector. To this end, the mainstream approaches make effort to introduce different spatial prior into DETR, which can be divided into implicit and explicit methods. Specifically, the former decouples reference coordinates from the learnable positional encodings, while the latter directly sets a 2D/4D coordinate for each query and maps such low-dimensional coordinate into a high dimension positional encoding via the sinusoidal PE~\cite{attention}. 

From the perspective of the spatial prior indoctrination, a straightforward way for object query is to predict the offset between their reference and the target bounding boxes. For example, previous approaches~
\cite{smca,conditionaldetr,wang2021anchor} only regress the offset of center points, while the current approaches~\cite{liu2022dab,zhang2022accelerating} directly regress the 4D offset based on the reference coordinate. Another spatial prior indoctrination benefits from the redesign of the cross-attention mechanism. Deformable DETR~\cite{deformabledetr}, SMCA~\cite{smca}, and SAM-DETR~\cite{zhang2022accelerating} aggregate multiple extreme  point regions from the content features by directly predicting the coordinates of these points from the object queries. Conditional DETR~\cite{conditionaldetr} and DAB-DETR~\cite{liu2022dab} utilize a Gaussian-like positional cross-attention map to attend to distinct regions dynamically. Take a close insight at the Gaussian map, the region of box sides and center point are attended by different heads in the multi-head attention mechanism. From the perspective of the spatial prior update, the prevailing approaches~\cite{liu2022dab,zhang2022accelerating} apply a cascaded way to refine the box prediction as well as update the reference spatial prior. \textit{However, all of these methods view center points as the reference spatial prior, eroding the discrimination of the positional encodings during performing the redundant prediction, thereby confusing the Transformer detector as well as leading to the slow model convergency.}

In our proposed SAP-DETR, such confusing reference spatial prior is replaced by the query-specific reference point. Specifically, each object query in SAP-DETR is assigned a non-overlapping fixed grid-region, which prompts queries to consider the grid area as a salient region to attend to image features and compensate for the over-smooth/inadequacy during center-based detection by localizing each side of the bounding box layer by layer. Considering the sparseness of the reference points, the movable strategy is proposed to enhance small/slender object detection. Therefore, there exists the 2D+4D reference spatial prior in the proposed SAP-DETR, and the final prediction is based on such a 6D reference coordinates ([$\Delta x,\Delta y,\Delta \, \Delta t,\Delta r, \Delta b $]). Taking an insight into the Conditional attention mechanism, we investigate that the highlight region is most relevant to four sides of bounding boxes, hence facilitating the final box localization. More intuitively, we devise the PECA to indicate the location of bounding box sides to object queries, where they should attend from context image features.

\section{Temperature Consistency in PE} 
\label{app:03}
\begin{figure}[t]
\centering
\includegraphics[width=3.3in]{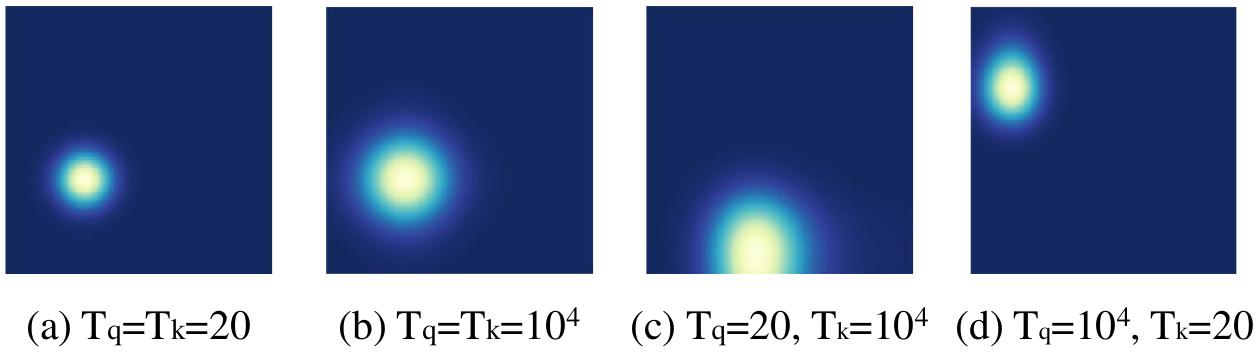}
\caption{Positional attention maps. Given two sequential PE of query-key pairs, we fix one PE of the query, reshape its sequential attention map for all PE of the key into original 2D image size.}
\label{fig:temp2}
\end{figure}

Following DETR, we also use the 2D sinusoidal function $\text{PE}(x,y)$ as positional encoding. Given a position, the $\text{PE}_\text{pos}$ is calculated by
\begin{equation}\label{eq:07}
    \begin{array}{l}
        \text{PE}_\text{pos}^T(i)=
        \begin{cases}
            \text{sin}(\text{pos} \cdot \omega_t)&  \quad i=2t\\
            \text{cos}(\text{pos} \cdot \omega_t)& \quad i=2t+1,\\
        \end{cases}
    \vspace{1ex}\\
     \quad \omega_t = T^{-2t/d}, \quad t=1,\cdots,d/2 ,
    \end{array}
\end{equation}
where $T$ is an adjustable temperature and $i$ is the channel index of the positional embedding. As shown in~\cref{fig:temp2}, the
\begin{figure}[t]
\centering
\includegraphics[width=3.3in]{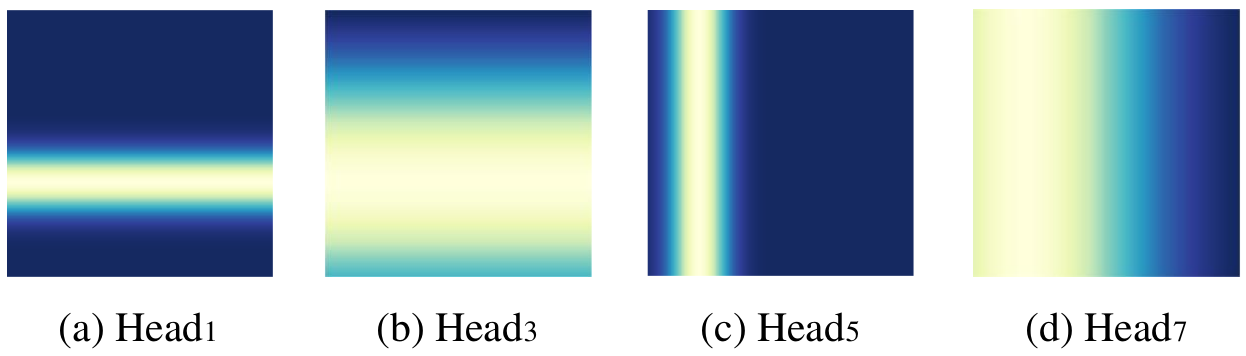}
\caption{Positional attention maps in each head. }
\label{fig:temp3}
\end{figure}

\noindent receptive field size of the positional attention map tends to become wider with increasing temperature~\cite{liu2022dab}. Before the softmax operation, the positional query-to-key similarity $\mathbf{A}$ in the cross-attention mechanism is computed by a dot-product between query position $\text{PE}_{\text{pos}_q}^{T_q}$ and key position $\text{PE}_{\text{pos}_k}^{T_k}$. Clearly, the resulting positional similarity in~\cref{fig:temp2}(a) and (b) subjects to a Gaussian-like distribution. We fix the $\text{PE}_{\text{pos}_q}^{T_q}$ and then the center of $\mathbf{A}$ is calculated by
\begin{equation}\label{eq:app01}
    \begin{aligned}
        \textbf{A}_{\text{PE}} &\!\!=\! \text{PE}_{\text{pos}_\text{q}}^{T_\text{q}} \cdot \; {\text{PE}_{\text{pos}_\text{k}}^{T_\text{k}}}^\top  \\
        &\!\!=\!\!\!\sum\limits_{t}^{d/2}\!\text{sin}(\omega_t^{\text{q}} \text{pos}_\text{q}\!)\text{sin}(\omega_t^{\text{k}} \text{pos}_\text{k}\!)\!+\!\text{cos}(\omega_t^{\text{q}} \text{pos}_\text{q}\!)\text{cos}(\omega_t^{\text{k}} \text{pos}_\text{k}\!)\\
        &\!\!=\!\!\! \sum\limits_{k}^{d/2}\!\text{cos}(\omega_t^{\text{q}} \text{pos}_\text{q}\!\!-\!\omega_t^{\text{k}} \text{pos}_\text{k}\!), \quad \text{pos}_\text{q}, \text{pos}_\text{k} \!\in\! \mathfrak{X} = [0, \dfrac{\pi}{2}].
    \end{aligned}
\end{equation}
By fixing the $\text{pos}_\text{q}$, the center $\text{pos}_\text{k}^{\text{center}(k)}$ of $\mathbf{A}_{\text{PE}}$ of each dimension $k\!\in\!\{1,\cdots\!,d/2\}$ is calculated by
\begin{equation}\label{eq:app02}
    \begin{aligned}
        \text{pos}_\text{k}^{\text{center}(t)}
        &=\mathop{\text{argmax}}\limits_{\text{pos}_\text{k} \in \mathfrak{X}}( \text{cos}(\omega_t^{\text{q}} \text{pos}_\text{q}-\omega_t^{\text{k}} \text{pos}_\text{k})) 
        \\ &=\mathop{\text{argmin}}\limits_{\text{pos}_\text{k} \in \mathfrak{X}}(\omega_t^{\text{q}} \text{pos}_\text{q}-\omega_t^{\text{k}} \text{pos}_\text{k})
        \\&=(T_\text{k}/T_\text{q})^{2t/d}\text{pos}_\text{q}.
    \end{aligned}
\end{equation}
Consequently, there exists an offset center for each channel of the positional attention map if $T_\text{k} \neq T_\text{q}$. Literally, each channel of the positional attention map can be viewed as a superposition by several horizontal and vertical line masks (see~\cref{fig:temp3}). So it is easy to illustrate the offset center and irregular width/height of the positional attention maps as shown in~\cref{fig:temp2}(c) and (d).

Without loss of generality, we eliminate the effect of conditional scaling transformation and fix the temperature of encoder's positional encoding to 20. As shown in~\cref{tab:peca}, the reported results compare the different temperature settings based on PECA. Clearly, both point and box site positional encodings are benefit from a relative small consistent temperature, especially when concatenating with box side PE.

\section{Scaling Transformation for PE} 
\label{app:02}
\noindent \textbf{Revisiting Conditional Spatial Query Prediction.}
Given a set of content queries and their corresponding reference points, the conditional spatial query prediction adaptively maps the reference points into high-dimensional positional embeddings according to a spatial transformation generated by content queries. Let $\boldsymbol{r}^\nshortparallel\!\in\!\mathbb{R}^{\text{k}}$ denotes the 2D unnormalized reference point, $\mathbf{e}\!\in\!\mathbb{R}^d$ denotes the content query, and $\mathbf{T}\!\in\!\mathbb{R}^d$ indexes the scaling spatial transformation where $d$ is the query dimension. Then the conditional spatial query prediction is calculated by
\begin{equation}\label{eq:app03}
    \begin{aligned}
    \mathbf{p}_\text{q}\!=\!\mathbf{T} \cdot \text{PE}(\text{sigmoid}(\boldsymbol{r}^\nshortparallel)), 
    \quad  
    \mathbf{T}\!=\!\text{FFN}(\mathbf{e}),
    \end{aligned}
\end{equation}
where FFN is a feed-forward network consisting of a linear layer, a ReLU activation, and a linear layer. PE is the sinusoidal positional encoding as illustrated in~\cref{eq:07}. In Conditional DETR~\cite{conditionaldetr}, the unnormalized reference point is either a learnable 2D coordinate or generated by its corresponding content query.

\noindent \textbf{Scaling Transformation in PECA.} 
As introduced in Section~\cref{app:03}, the proposed PECA concatenates both point and box side PEs for conditional spatial cross-attention. Following the scaling transformation of Conditional DETR, we also conduct ablations on different ways of scaling transformation in PECA. The following settings are involved:
\begin{itemize}
\item Comparing the effectiveness of scaling transformation with and without box side PE concatenation.
\item Comparing the effectiveness of scaling consistency in both point PE and box side PE, and then considering three types of ablation: no scaling, shared, and independent scaling transformation.
\item Exploiting a learnable diagonal matrix to transform the positional encoding of the key-vector, which also can be shared between point PE and box side PE. 
\end{itemize}
\cref{tab:scalingtranfer} summarizes the results of the ablation study on the 3-layer encoder-decoder Transformer neck. There exists a large gap between the performances of the model with no key-vector scaling transformation and counterparts with the transformation. We speculate that the scaling transformation of key-vector PE may cause the decoder confusion in extreme region localization, while the transformation on query-vector PE (point or box side) would facilitate it to focus on the spatial information within the content embeddings to the content image features. In addition, we observe that the main function of the point PE is to keep reference-specific for each query, and its effectiveness on box side attention will be weakened when concatenating the box side PE. Finally, we use a shared scaling transformation for both point and box side PEs.
\begin{table*}[t]
  \centering
    \setlength \tabcolsep{10pt}
    \begin{spacing}{0.7}
    \centering
    \resizebox{0.78\linewidth}{!}{%
    \begin{tabular}{cccccccccc}
    \toprule
    \multirow{2}[2]{*}{\makecell[c]{Concatenate \\ Box Side PE}}&\multicolumn{3}{c}{Scaling Transformation for PE} & \multicolumn{1}{c}{\multirow{2}[2]{*}{AP}} & \multicolumn{1}{c}{\multirow{2}[2]{*}{AP$_{\text{50}}$}}& \multicolumn{1}{c}{\multirow{2}[2]{*}{AP$_{\text{75}}$}} & \multicolumn{1}{c}{\multirow{2}[2]{*}{AP$_{\text{S}}$}} & 
    \multicolumn{1}{c}{\multirow{2}[2]{*}{AP$_{\text{M}}$}} &
    \multicolumn{1}{c}{\multirow{2}[2]{*}{AP$_{\text{L}}$}} \\
    \cmidrule{2-4}  
    &\quad\ $\mathbf{T}_k$\ \quad & \quad\ $\mathbf{T}_{qp}$\ \quad & $\mathbf{T}_{qb}$ &       &&&       & & \\
    \toprule
    \ding{55}& \quad\ \tiny{$\Circle$}\ \quad  & \quad\ \tiny{$\Circle$}\ \quad &   -  & 33.3 & 54.4 & 33.9 & 13.1 & 36.3 & 52.2 \\
    \ding{55}& \quad\ \tiny{$\CIRCLE$}\ \quad  & \quad\ \tiny{$\Circle$}\ \quad &   -  & 32.3 & 53.4 & 32.7 & 12.8 & 35.4 & 50.4 \\
    \ding{55}& \quad\ \tiny{$\Circle$}\ \quad  & \quad\ \tiny{$\CIRCLE$}\ \quad &   -  & 34.4 & 55.3 & 35.6 & 14.5 & 37.7 & 53.3 \\
    \ding{55}& \quad\ \tiny{$\CIRCLE$}\ \quad  & \quad\ \tiny{$\CIRCLE$}\ \quad &   -  & 32.6 & 53.7 & 33.1 & 12.2 & 35.5 & 51.1 \\
    \specialrule{0.5pt}{0.7pt}{1.5pt}
    \checkmark& \quad\ \tiny{$\Circle$}\ \quad & \quad\ \tiny{$\Circle$}\ \quad & \tiny{$\Circle$} & 34.0 & 54.5 & 35.0 & 13.9 & 37.1 & 52.8 \\
    \checkmark& \quad\ \tiny{$\CIRCLE$}\ \quad & \quad\ \tiny{$\Circle$}\ \quad & \tiny{$\Circle$} & 33.2 & 54.0 & 34.0 & 13.3 & 36.4 & 51.7 \\
    \checkmark& \quad\ \tiny{$\Circle$}\ \quad & \quad\ \tiny{$\CIRCLE$}\ \quad & \tiny{$\Circle$} & 34.7 & 55.3 & 35.8 & 14.4 & 37.8 & 53.1 \\
    \checkmark& \quad\ \tiny{$\Circle$}\ \quad & \quad\ \tiny{$\Circle$}\ \quad & \tiny{$\CIRCLE$} & 35.1 & 55.1 & 36.7 & 14.9 & 38.2 & 53.5 \\
    \checkmark& \quad\ \tiny{$\CIRCLE$}\ \quad & \quad\ \tiny{$\CIRCLE$}\ \quad & \tiny{$\CIRCLE$} & 32.6 & 53.4 & 33.2 & 12.3 & 35.6 & 52.0 \\
    \checkmark& \quad\ \tiny{$\Circle$}\ \quad & \quad\ \tiny{$\LEFTcircle$}\ \quad & \tiny{$\RIGHTcircle$} & 35.2 & 55.1 & 36.6 & 15.7 & 38.5 & \textbf{53.9 }\\
    \checkmark& \quad\ \tiny{$\Circle$}\ \quad & \quad\ \tiny{$\CIRCLE$}\ \quad & \tiny{$\CIRCLE$} & \textbf{35.2} & \textbf{55.4} & \textbf{36.8} & \textbf{15.8} & \textbf{38.5} & 53.6 \\
    \bottomrule
    \specialrule{0pt}{0pt}{1.5pt}
    \multicolumn{10}{l}{\tiny{$^{\LEFTcircle}$}\!\!\: \normalsize{and} \tiny{$^{\RIGHTcircle}$} \normalsize{denote different independent scaling transformations.}} \\
    \multicolumn{10}{l}{\tiny{$^{\Circle}$} \normalsize{and} \tiny{$^{\CIRCLE}$} \normalsize{denote no scaling and shared scaling transformations, respectively.}} \\
    \bottomrule
    \end{tabular}}
    \end{spacing}
    \caption{Ablation study on the scaling transformation of PE.}
    \vspace{0.25cm}
    \label{tab:scalingtranfer}%
\end{table*}
\begin{table*}[t]
  \centering
   \setlength \tabcolsep{10pt}
    \begin{spacing}{0.7}
    \resizebox{0.78\linewidth}{!}{%
    \begin{tabular}{cccccccccc}
    \toprule
    \multirow{2}[2]{*}{\makecell[c]{Concatenate \\ Box Side PE}}&\multicolumn{3}{c}{Temperature of PE} & \multicolumn{1}{c}{\multirow{2}[2]{*}{AP}} &\multicolumn{1}{c}{\multirow{2}[2]{*}{AP$_\text{50}$}} &\multicolumn{1}{c}{\multirow{2}[2]{*}{AP$_\text{75}$}} & \multicolumn{1}{c}{\multirow{2}[2]{*}{AP$_\text{S}$}} & 
    \multicolumn{1}{c}{\multirow{2}[2]{*}{AP$_\text{M}$}} &
    \multicolumn{1}{c}{\multirow{2}[2]{*}{AP$_\text{L}$}} \\
    \cmidrule{2-4}  
    &$T_k$ & $T_{qp}$ & $T_{qb}$ &       &       & & \\
    \toprule
    \ding{55}&         $20$  &     $1000$&   -         & 31.8 & 52.8 & 32.1 & 12.8 & 34.4 & 50.5 \\
    \ding{55}&        $1000$ &       $20$&   -         & 32.1 & 53.0 & 32.5 & 12.7 & 35.2 & 50.6 \\
    \ding{55}&           $20$&       $20$&   -         & 32.2 & 53.2 & 32.7 & 12.7 & 35.0 & 51.0 \\ 
    \specialrule{0.5pt}{0.7pt}{1.5pt}
    \checkmark&         $20$&       $1000$&      $1000$& 32.2 & 52.9 & 32.9 & 12.8 & 35.1 & 51.1 \\
    \checkmark&       $1000$&         $20$&        $20$& 32.3 & 52.7 & 32.8 & 13.3 & 35.2 & 50.8 \\
    \checkmark&         $20$&         $20$&        $20$& \textbf{33.0} & \textbf{53.6} & \textbf{33.5} & \textbf{13.7} & \textbf{36.3} & \textbf{52.1} \\
    \bottomrule
    \end{tabular}}
  \end{spacing}
  \caption{Ablation Study on the temperature consistency of PE.}
  \label{tab:peca}%
\end{table*}%

\begin{figure}
\centering
\includegraphics[width=3.2in]{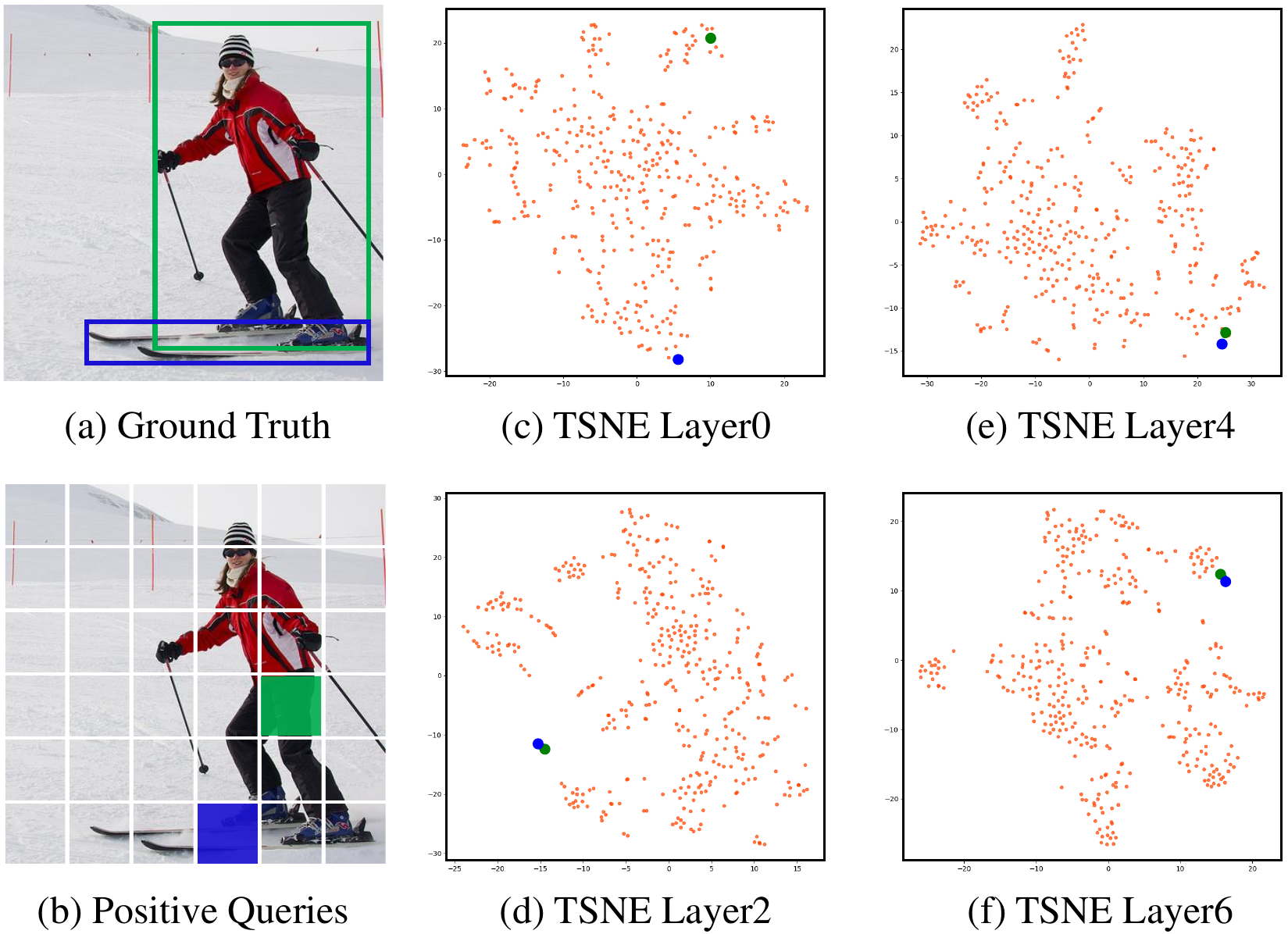}
\caption{Visualization of t-SNE. Both grids and slots in t-SNE represent object queries, where the green and blue color are the positive queries, corresponding to the same colored ground truth.}
\label{fig:tsne}
\end{figure}%


\begin{table*}[t]
  \centering
   \setlength \tabcolsep{12pt}
    \begin{spacing}{0.7}
    \resizebox{0.78\linewidth}{!}{%
    \begin{tabular}{ccccccccc}
    \toprule
    \multirow{2}[2]{*}{Detach}&\multicolumn{2}{c}{Indep. Prediction Head } & \multicolumn{1}{c}{\multirow{2}[2]{*}{AP}} & \multicolumn{1}{c}{\multirow{2}[2]{*}{AP$_\text{50}$}}& \multicolumn{1}{c}{\multirow{2}[2]{*}{AP$_\text{75}$}}& \multicolumn{1}{c}{\multirow{2}[2]{*}{AP$_\text{S}$}} & 
    \multicolumn{1}{c}{\multirow{2}[2]{*}{AP$_\text{M}$}} &
    \multicolumn{1}{c}{\multirow{2}[2]{*}{AP$_\text{L}$}} \\
    \cmidrule{2-3}  
    &\quad Head$_{\text{cls}}$ & Head$_{\text{bbox}}$  &&&       &       & & \\
    \toprule
    \ding{55}&             &                  &     34.6      & 54.8 & 35.7 & 14.4 & 37.6 & 52.8 \\
    \ding{55}&   \checkmark&                  &     34.7      & 54.8 & 36.0 & \textbf{16.2} & 37.6 &  53.2 \\
    \ding{55}&             &   \checkmark     &     34.8      & 55.0 & 35.8 & 15.3 & 37.8 & 53.5 \\
    \ding{55}&   \checkmark&    \checkmark    &     34.7      & 54.8 & 36.1 & 14.5 & 38.1 & 52.7 \\
    \specialrule{0.5pt}{0.7pt}{1.5pt}
    \checkmark&            &                  &     35.0      & 55.1 & 36.5 & 15.6 & 38.3 & 53.0 \\
    \checkmark& \checkmark &                  &     34.6      & 55.1 & 35.9 & 14.6 & 37.9 & 52.1 \\
    \checkmark&            & \checkmark       & \textbf{35.2} & \textbf{55.4} & \textbf{36.8} & 15.8 &  \textbf{38.5} &  53.6 \\
    \checkmark& \checkmark &  \checkmark      &     35.0      & 55.2 & 36.3 & 14.7 & 38.2 & \textbf{54.0} \\
    \bottomrule
    \end{tabular}}
  \end{spacing}
  \caption{Ablation study on the independent prediction head.}
  \label{tab:diffhead}%
\end{table*}%

\section{Independent Prediction Heads} 
\label{app:04}
Taking a close insight into the semantic representation of these object queries, we map each query output into a 2D distribution via t-SNE~\cite{van2008visualizing}. As shown in~\cref{fig:tsne}, each dot here represents an query output from the decoder layer. It can be seen that the instance objects (blue and green dots in~\cref{fig:tsne}(c)-(f)) whose location at the edge/corner of the distribution are easy to distinguish from the background queries. More precisely, the instance objects, except from the first decoder layer, are at a closer distance than the semantic-close queries. Inspired by this, we employ a dedicated classification head for the first decoder layer and a shared head for the others in the auxiliary training process.

\cref{tab:diffhead} reports the ablation study on the 3-layer encoder-decoder decoder neck. As we can see, the detach operation generally boosts the detector performance by $\sim$0.3\%AP, and the independent box prediction head is conducive to the Transformer detector for further improvements. Moreover, There exists a slight performance drop when using the independent classification prediction head.

\section{Movable Reference Points} 
\label{app:05}
\begin{figure*}[ht]
\centering
\includegraphics[width=0.73\linewidth]{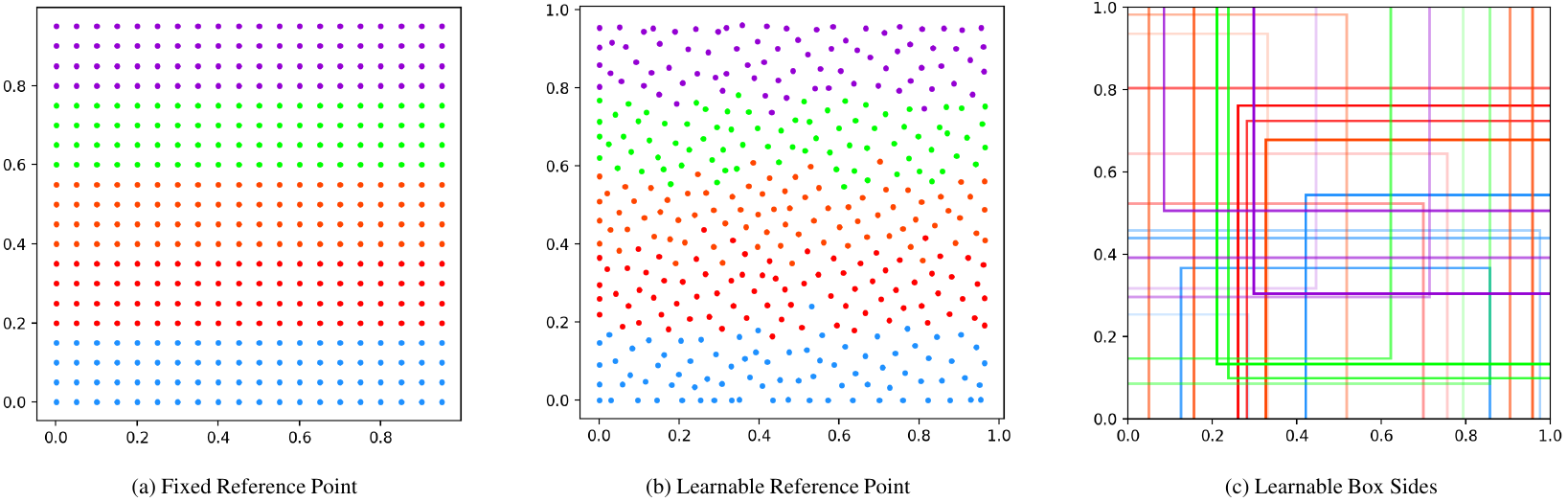}
\caption{Distribution.}
\vspace{0.2cm}
\label{fig:reference_distribution}
\end{figure*}

We evaluate two types of training strategies for reference points. As illustrated in~\cref{fig:reference_distribution}(a), we tile the mesh-grid reference points for their initialization and set such coordinates as fixed/learnable parameters. By visualizing the learnable reference points in~\cref{fig:reference_distribution}(b), their distribution are observed to be uniform within the image, similar to the learnable anchor points in Anchor DETR~\cite{wang2021anchor}. It indicates that the learnable reference coordinates would not be affected by properties of the target regression. We further hypothesize that there exists partial denominators between salient points and the center anchor points, to a certain extent. 

\begin{figure*}[ht]
\centering
\begin{subfigure}{0.46\linewidth}
    \centering
    \includegraphics[width=1\linewidth]{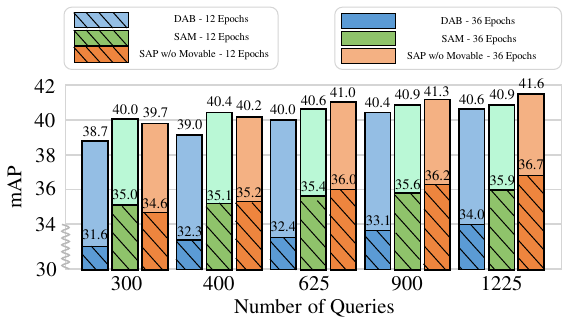}
    \caption{w/o Movable Reference Point}
\end{subfigure}
\begin{subfigure}{0.46\linewidth}
    \centering
    \includegraphics[width=1\linewidth]{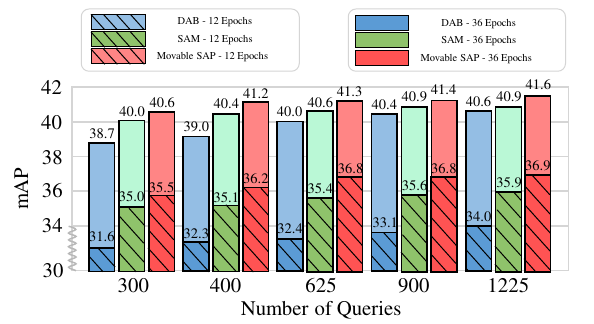}
    \caption{Movable Reference Point}
\end{subfigure}    
\caption{Comparison of performance and training losses curves between our purposed SAP-DETR and the current SOTA methods.}
\label{fig:movable_querynum_growth}
\end{figure*}

As introduced in~\cref{sec:031}, the proposed movable reference points significantly facilitate detecting small and slender objects, which are omitted caused by the sparseness of the reference point distribution. The experiments in~\cref{sec:042} demonstrate that the performance of small object detection is prompted after applying the movable strategy. Dialectically, we conduct another ablation on the number of queries to verify that such a vulnerability is attribute to the query sparsity. \cref{fig:movable_querynum_growth} describes the performance histogram of 3-layer detectors based on both 12-epoch and 36-epoch training schemes. Without the help of the movable component, the standard SAP-DETR relatively benefits more from the query number growth compared to the counterpart. Along with query number increase, the performance gap is reduced progressively (from 1.2 AP to 0.2 AP), which further verifies our sparsity analysis and the effectiveness of the movable strategy.

\begin{figure*}[htb]
\centering
\vspace{-1cm}
\includegraphics[width=0.77\linewidth]{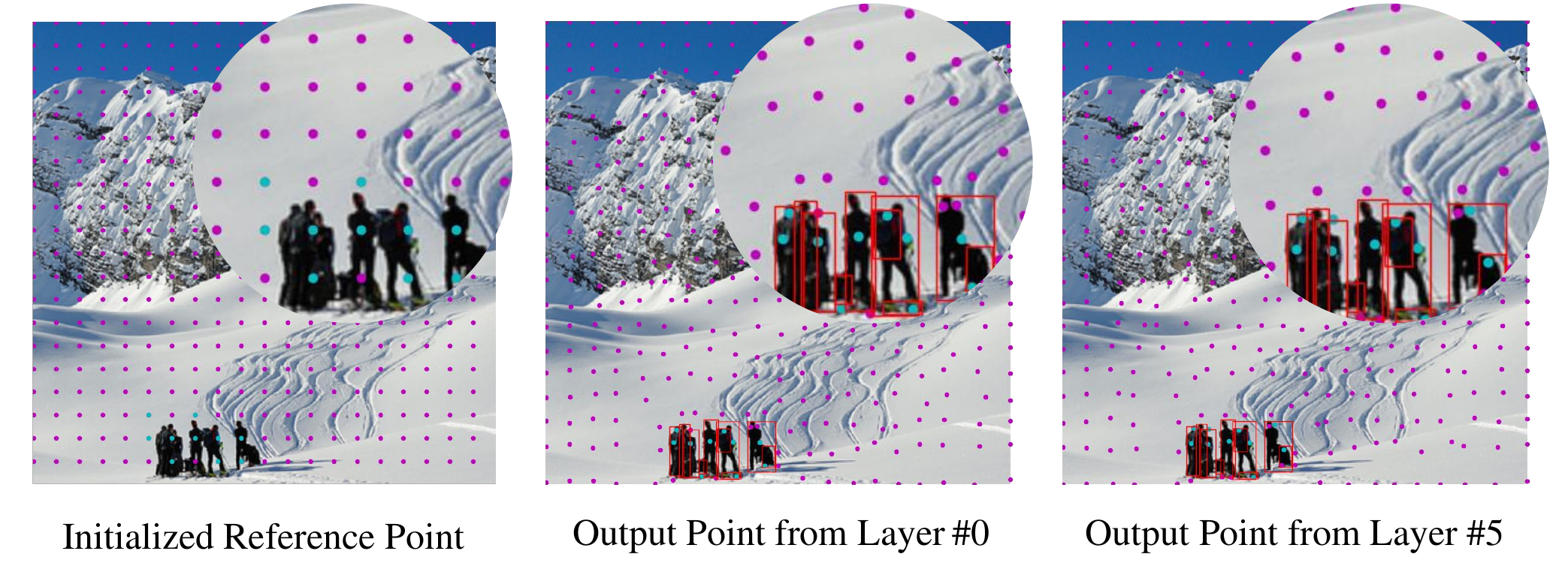}
\caption{Movable point update for COCO validation image \#3255.}
\label{fig:movable_point_update_3255}
\end{figure*}

\begin{figure*}[htb]
\centering
\includegraphics[width=0.77\linewidth]{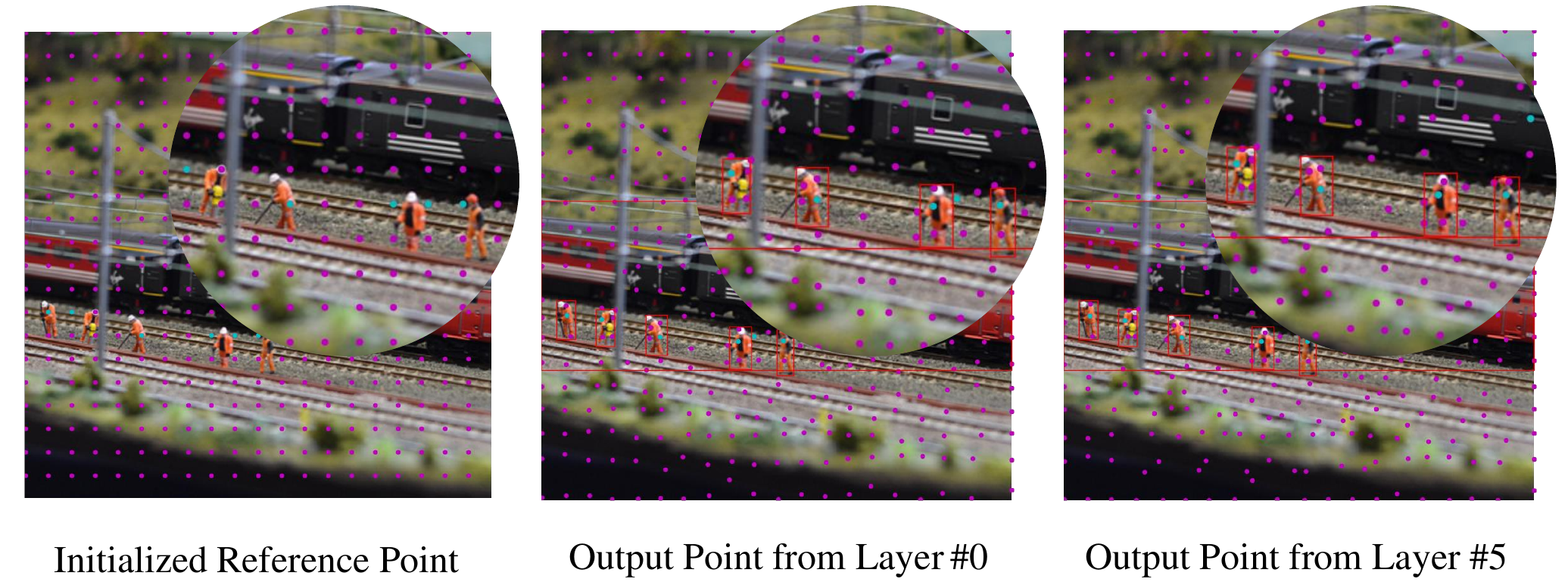}
\caption{Movable point update for COCO validation image \#14473.}
\label{fig:movable_point_update_14473}
\end{figure*}

To further demonstrate the effectiveness of the movable strategy, the update processes of salient points are plotted in~\cref{fig:movable_point_update_3255} and~\cref{fig:movable_point_update_14473}. Indeed, some small and slender objects can be localized well after moving the reference point within the objects. However, some queries whose reference points are located within the large objects behave an unstable matching result that the matched queries in the latter layers are inconsistent with the previous layers. Hence there exists a slight performance deterioration for large object detection after adding the movable reference points.

\section{Training Details and More Configurations} 
\label{app:07}
\noindent \textbf{Warm Up Training Strategy.} In the early training process, the bipartite matching in Transformer detectors may appear to be fragile and instable, where the positive label are assigned to one false prediction. This phenomenon is also reported in DN-DETR~\cite{li2022dn}. Following the conventional training strategy, we conduct a warm-up step during the early training process. In our experiments, we set warm-up steps to 400 and 1000 iterations for 3-layer and 6-layer Encoder-Decoder Transformer detectors, respectively.

\noindent \textbf{Detailed Configurations.}
We list the all configurations in~\cref{tab:configuration}. For each number of query in~\cref{app:05}, the batch size of 8 is applied in our 3-layer SAP-DETR. 

\section{Visualization of Attention Maps} 
\label{app:08}
\begin{figure*}[ht]
    \centering
	\subfloat[Ground Truth of COCO validation image \#785]{
	    \centering
        \includegraphics[width=0.52\linewidth]{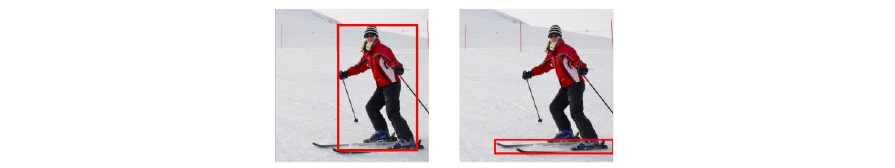}
    } \\ \vspace{0.4cm}
    \subfloat[DAB-DETR for COCO validation image \#785]{
        \centering
        \includegraphics[width=0.78\linewidth]{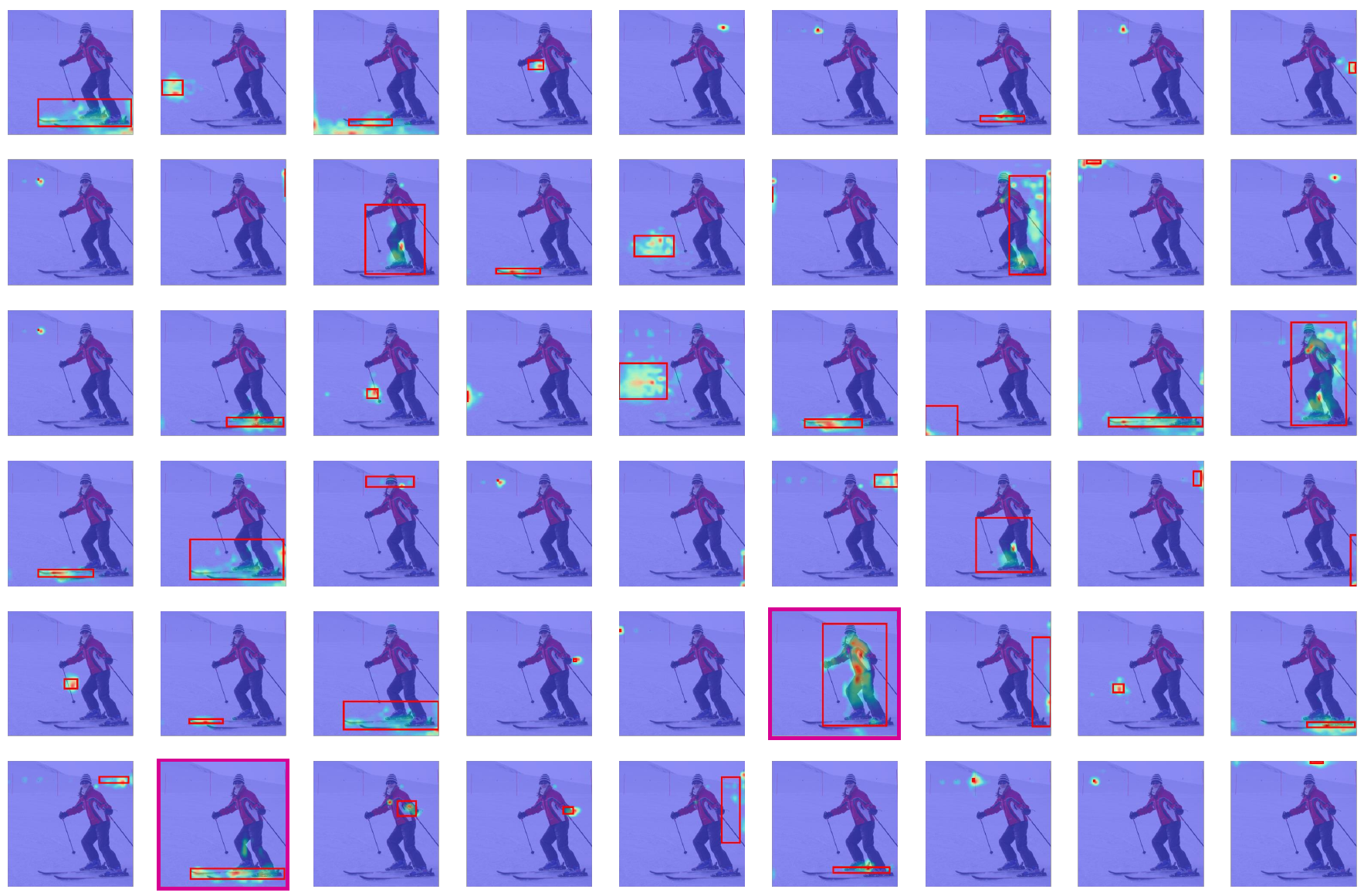}
    } \\ \vspace{0.4cm}
	\subfloat[SAP-DETR for COCO validation image \#785]{
	    \centering
        \includegraphics[width=0.78\linewidth]{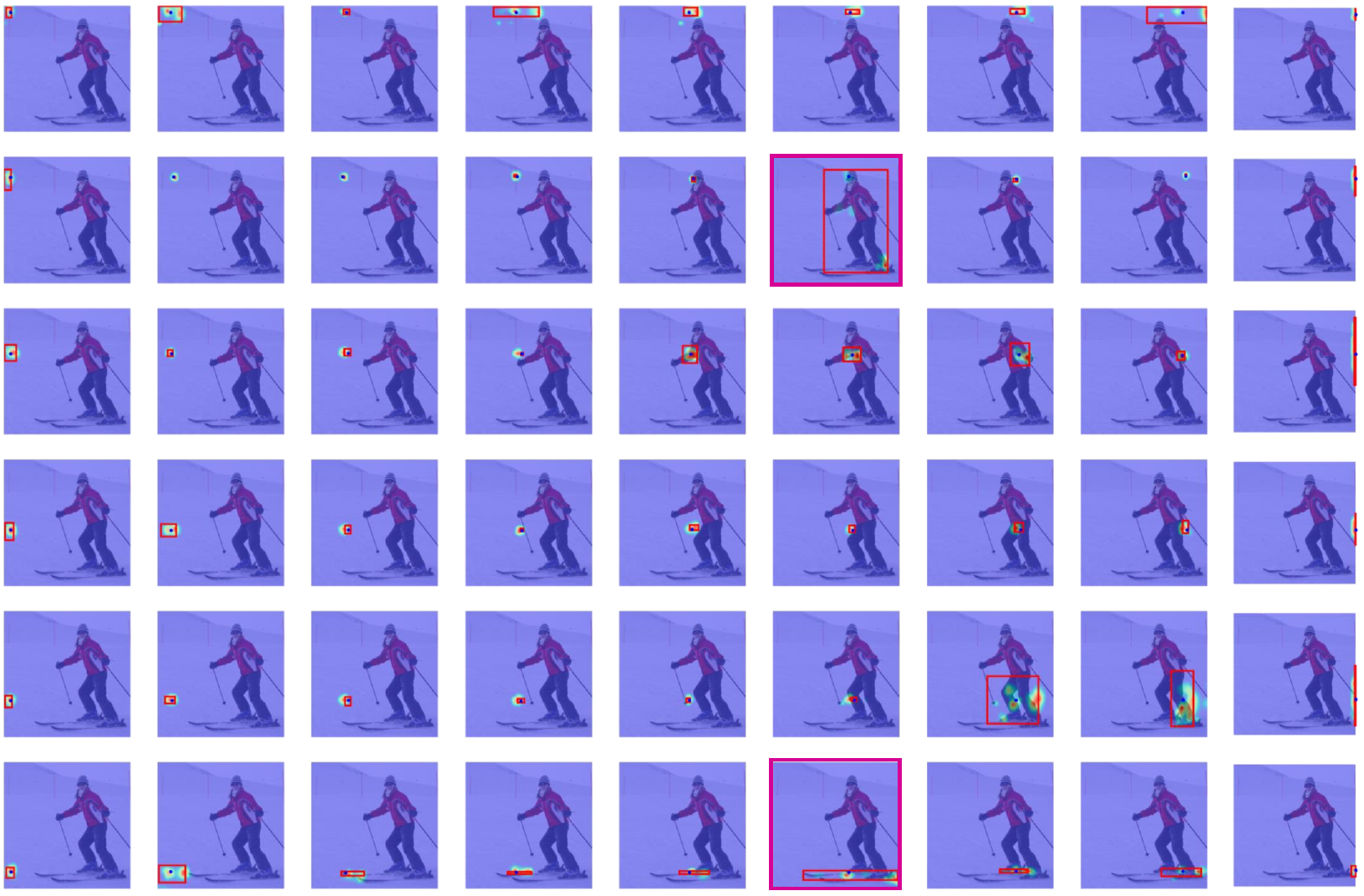}
    }
    \caption{Visualization of partial object queries in both SAP-DETR and DAB-DETR.}
    \label{fig:dab_meshgrid_attn_785}
\end{figure*}

\begin{figure*}[ht]
    \centering
	\subfloat[Ground Truth of COCO validation image \#71226]{
    \includegraphics[width=0.5\linewidth]{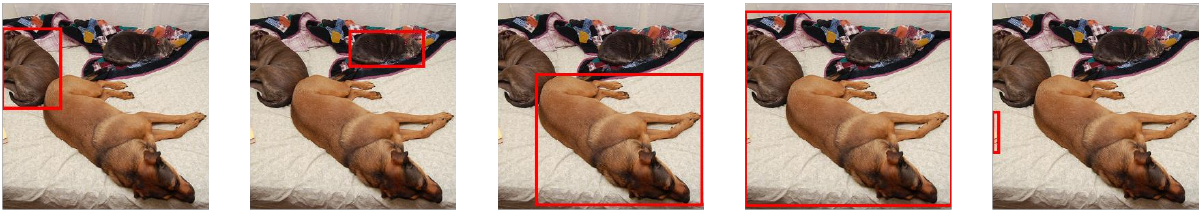}
    }\\ \vspace{0.4cm}
	\subfloat[DAB-DETR for COCO validation image \#71226]{
    \includegraphics[width=0.81\linewidth]{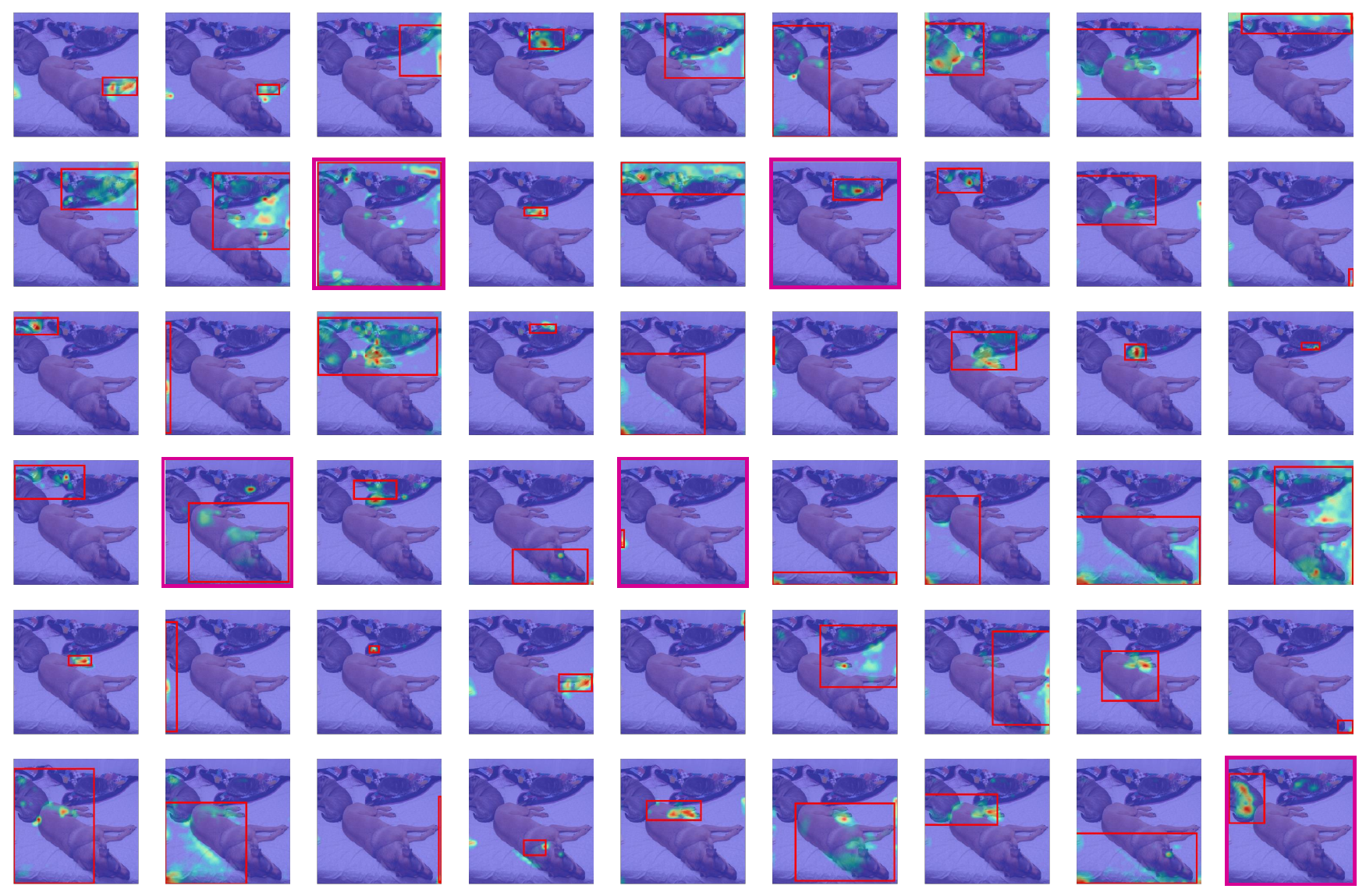}
    }\\ \vspace{0.4cm}
	\subfloat[SAP-DETR for COCO validation image \#71226]{
    \includegraphics[width=0.81\linewidth]{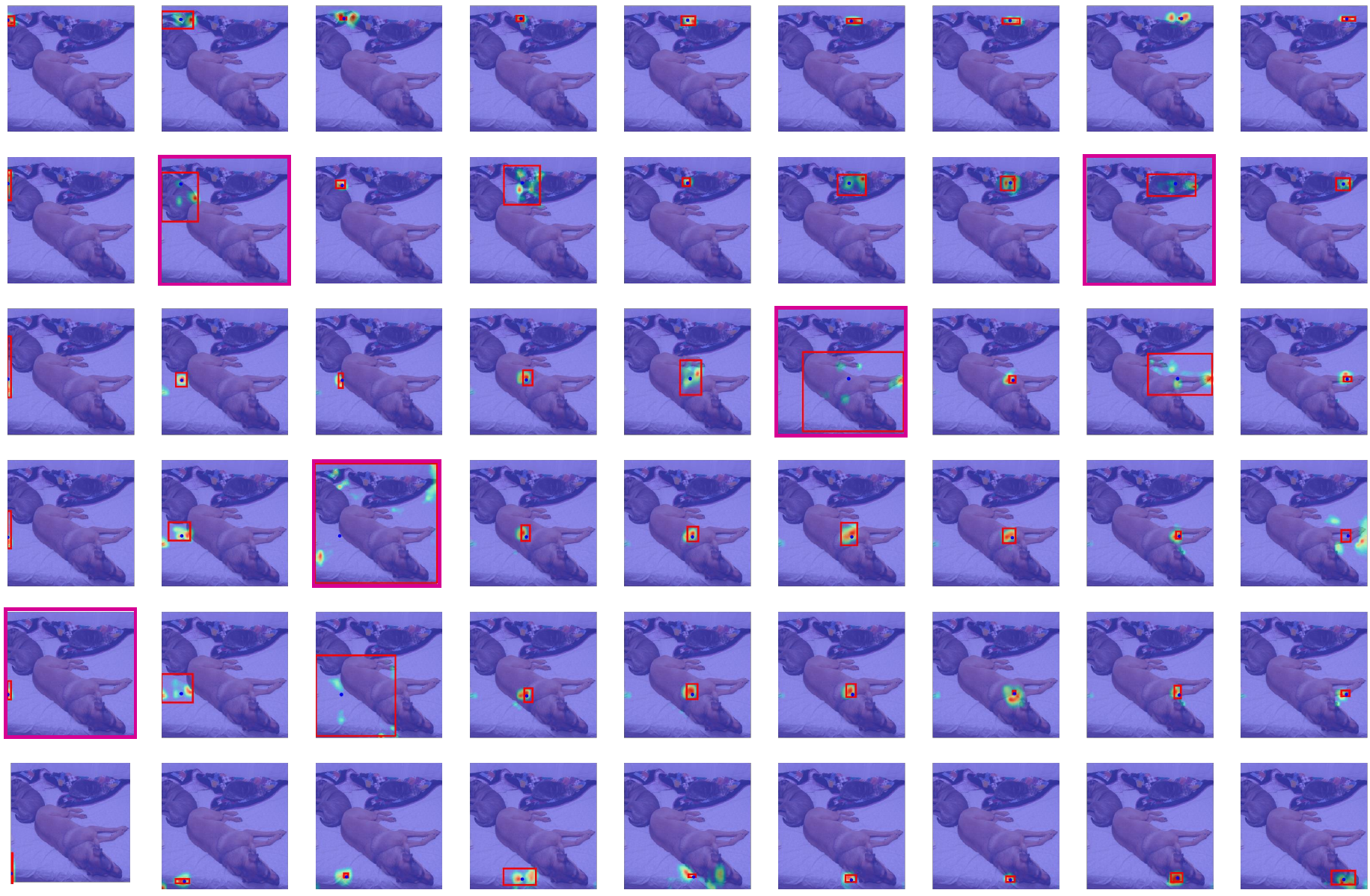}
    }
    \caption{Visualization of partial object queries in both SAP-DETR and DAB-DETR.}
    \label{fig:dab_meshgrid_attn_71226}
\end{figure*}

\begin{figure*}[ht]
    \centering
	\subfloat[Ground Truth of COCO validation image \#1000]{
    \includegraphics[width=0.55\linewidth]{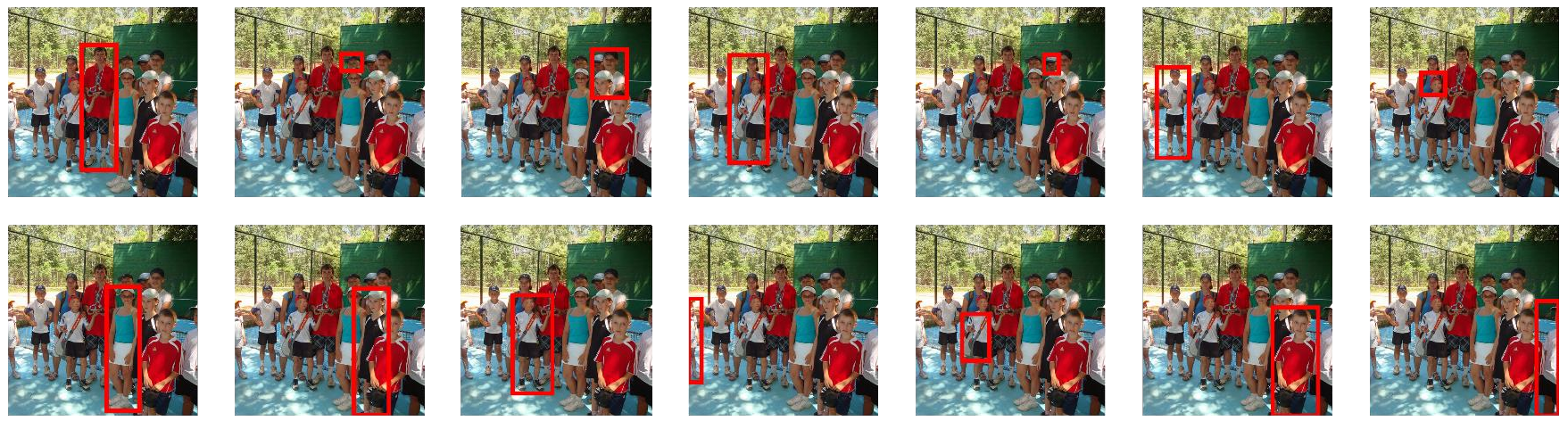}
    }\\ \vspace{0.4cm}
	\subfloat[DAB-DETR for COCO validation image \#1000]{
    \includegraphics[width=0.8\linewidth]{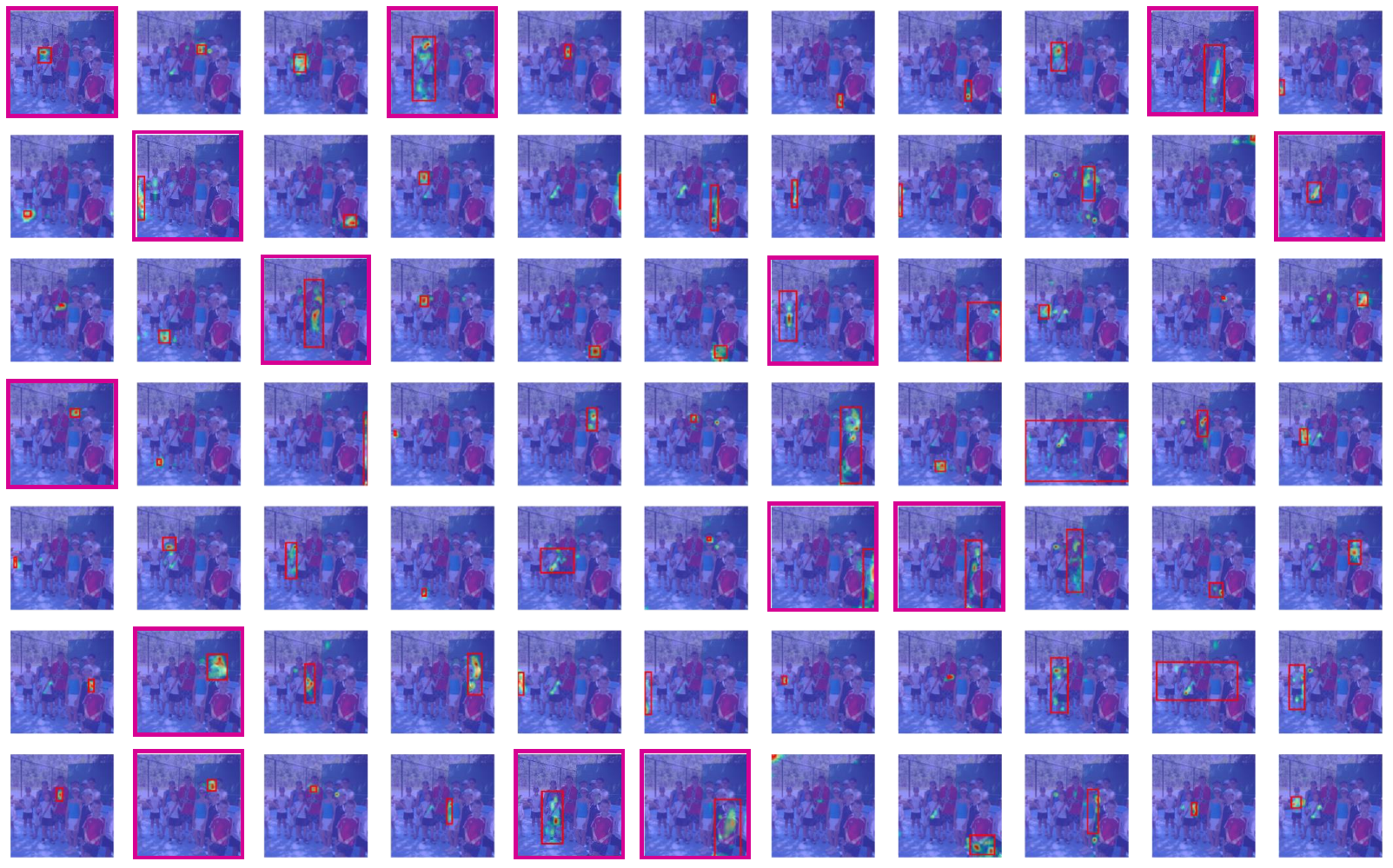}
    }\\ \vspace{0.4cm}
	\subfloat[SAP-DETR for COCO validation image \#1000]{
    \includegraphics[width=0.8\linewidth]{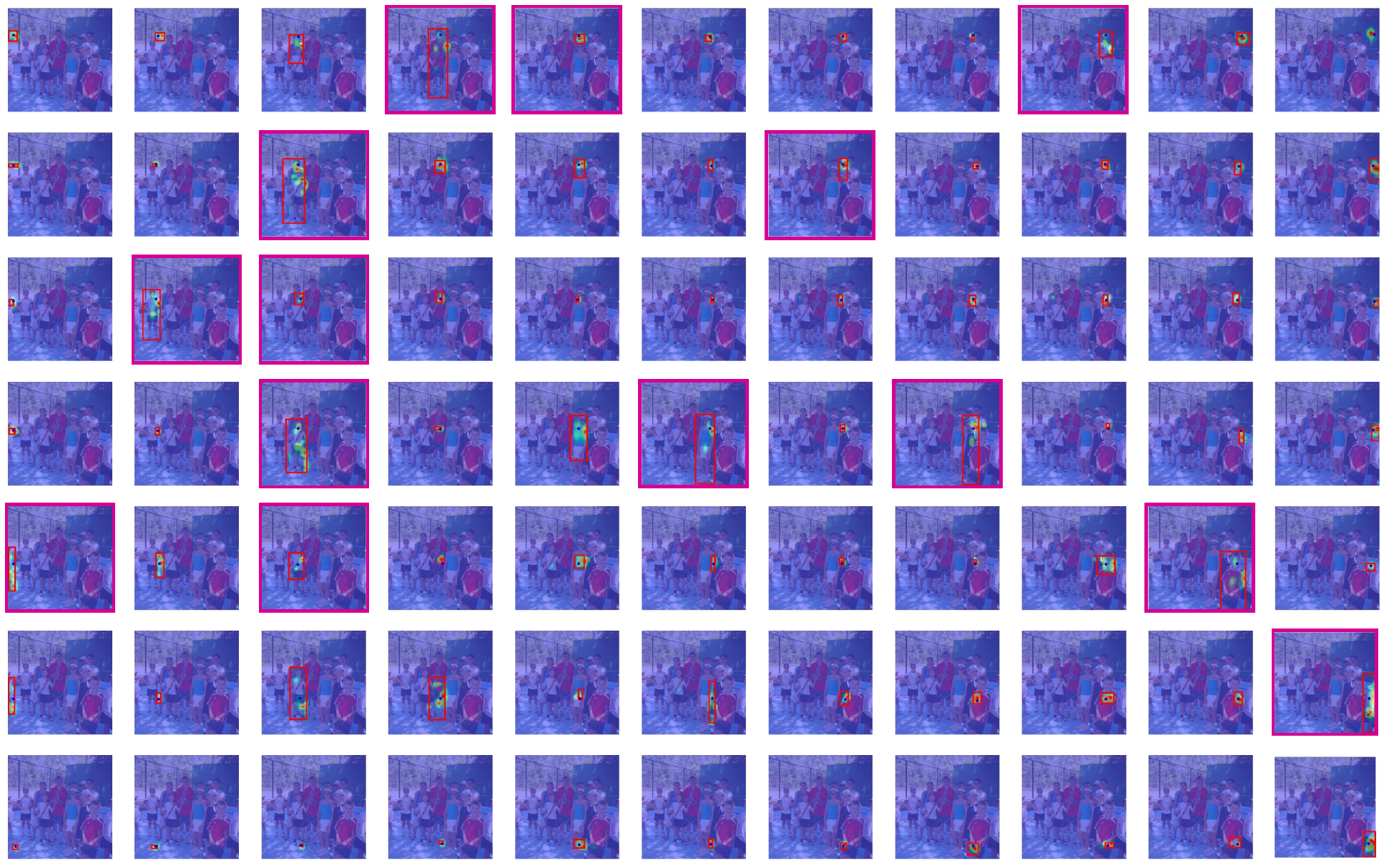}
    }
    \caption{Visualization of partial object queries in both SAP-DETR and DAB-DETR.}
    \label{fig:dab_meshgrid_attn_1000}
\end{figure*}

\begin{figure*}[ht]
    \centering
	\subfloat[Ground Truth of COCO validation image \#3225]{
    \includegraphics[width=0.45\linewidth]{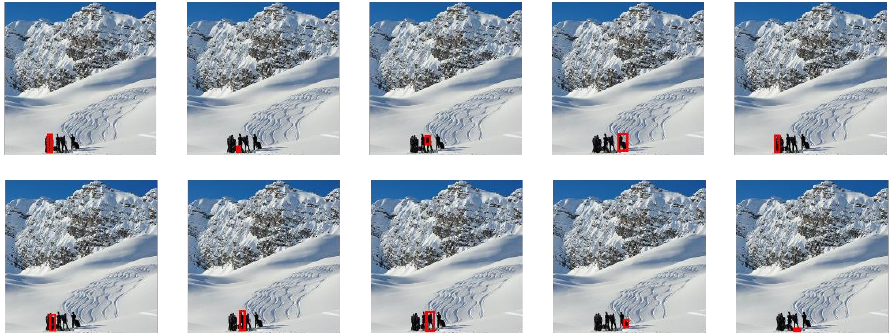}
    }\\ \vspace{0.4cm}
	\subfloat[DAB-DETR for COCO validation image \#3255]{
    \includegraphics[width=0.76\linewidth]{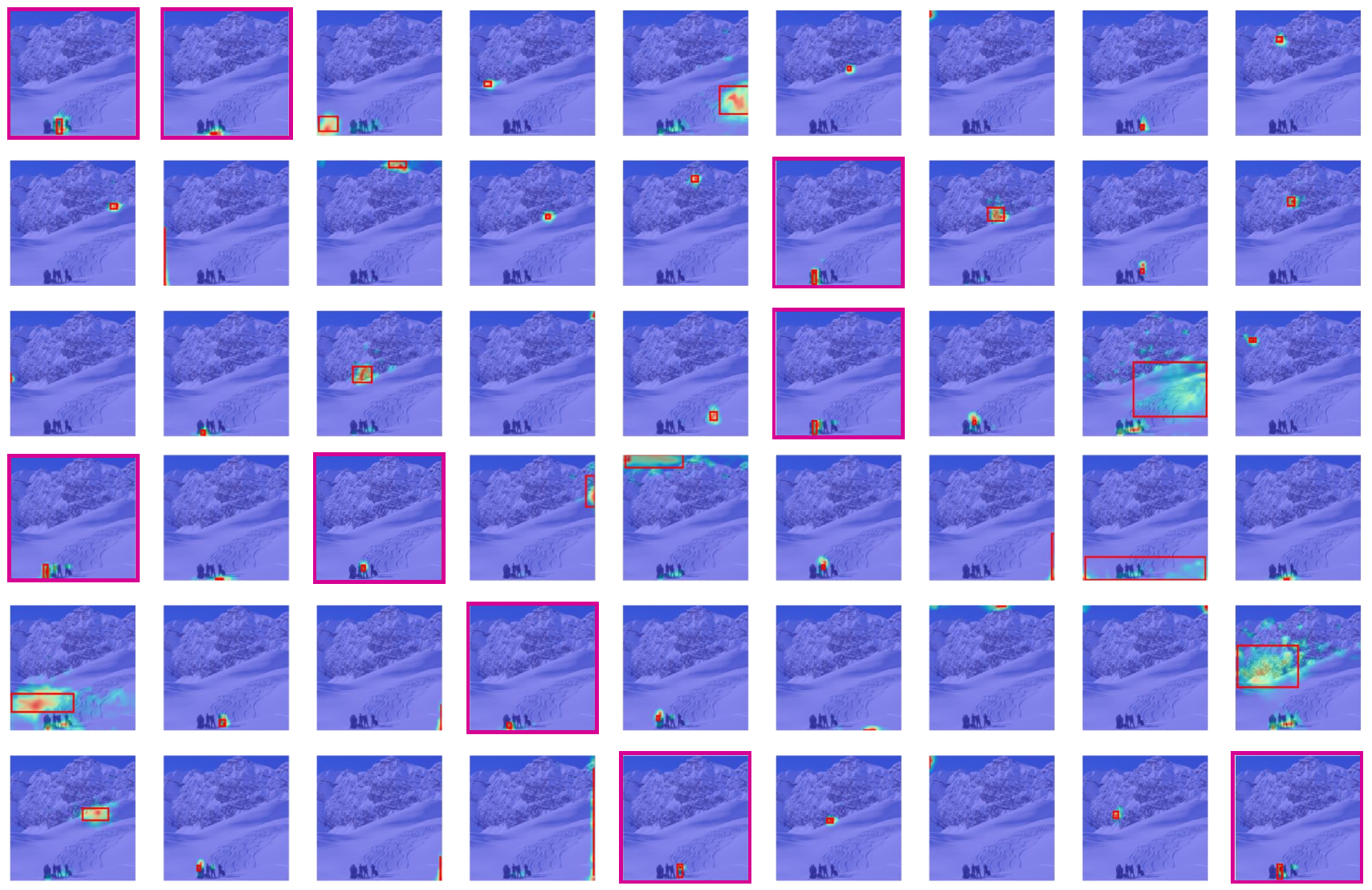}
    } \\ \vspace{0.4cm}
	\subfloat[SAP-DETR for COCO validation image \#3255]{
    \includegraphics[width=0.76\linewidth]{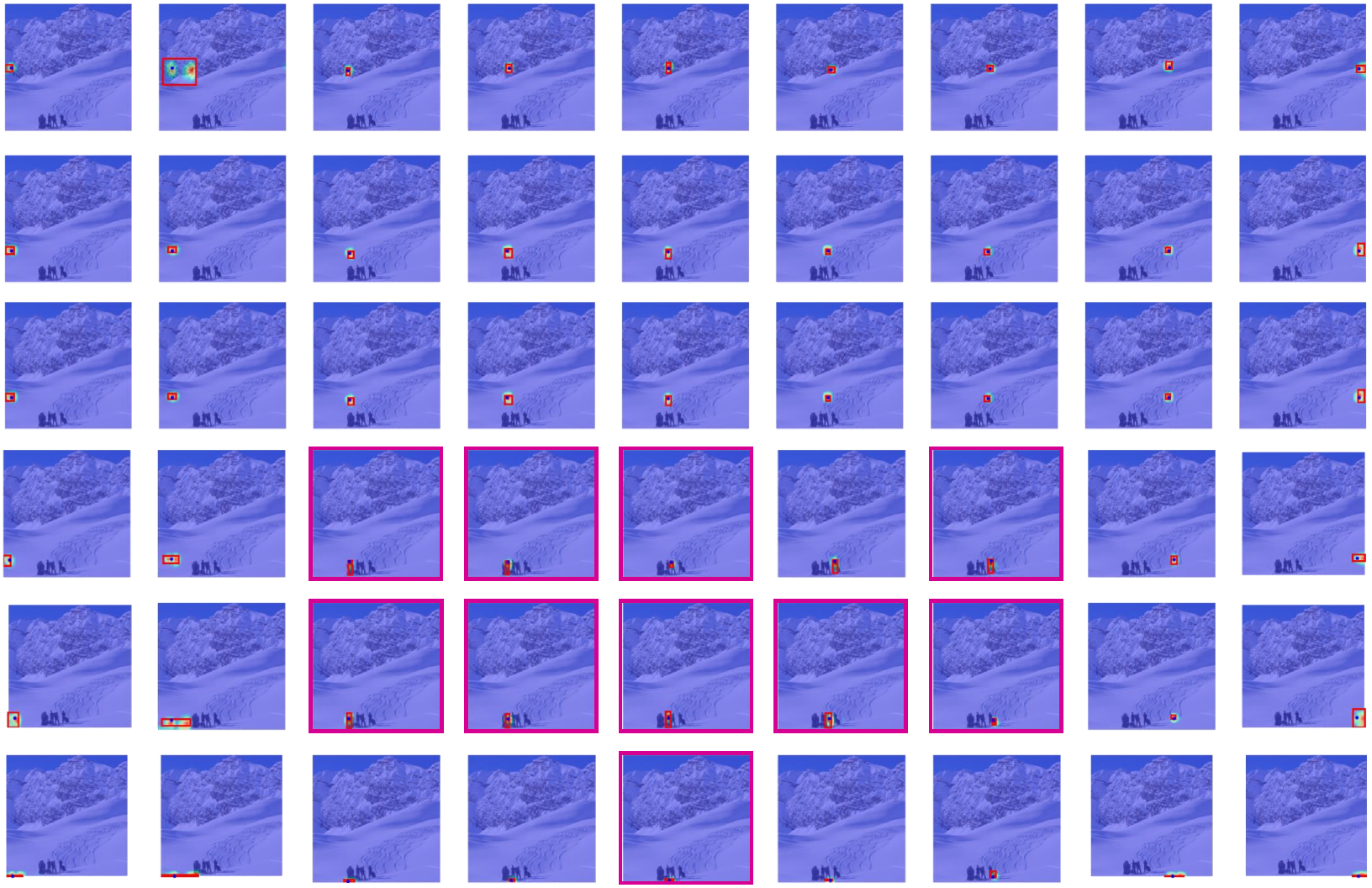}
    }
    \caption{Visualization of partial object queries in both SAP-DETR and DAB-DETR.}
    \label{fig:dab_meshgrid_attn_3225}
\end{figure*}

\noindent \textbf{Visualization of Query-Specific Region.} To understand how query-specific reference point affect on the object queries aggregation, we visualize the cross-attention map and the output bounding box for each query based on DAB-DETR and our proposed SAP-DETR in~\cref{fig:dab_meshgrid_attn_785} to~\cref{fig:dab_meshgrid_attn_3225}. 
Precisely, we visualize the query-specific region in various scenes. For example, the \#785 validation image with sample background and sparse instance, the \#71226 validation image with complex background and different scale objects, the \#1000 validation image with sophisticated instance objects, and the \#3255 validation image with sophisticated small instance objects. Compared with redundant prediction and wilderness attention region in DAB-DETR, each query of SAP-DETR only has a compact attention receptive field except for the positive instance query, which benefits from the query-specific reference point and PECA attention mechansim, hence resulting in a superior convergency speed.

\begin{table}[t]
  \centering
  \vspace{0.2cm}
  \setlength \tabcolsep{5pt}
    \begin{spacing}{0.85}
    \resizebox{0.491\linewidth}{!}{%
    \begin{tabular}{lc}
    \toprule
    Item  & Value \\
    \midrule
    lr    & 1e-4\\
    lr\_backbone & 1e-5 \\
    weight\_decay & 1e-4 \\
    k\_pe\_temp & 20 \\ 
    q\_point\_pe\_temp & 20 \\
    q\_bbox\_pe\_temp & 20 \\
    enc\_layers & 3 / 6 \\
    dec\_layers & 3 / 6   \\
    dim\_feedforward \hspace{0.66cm} & 2048  \\
    hidden\_dim & 256   \\
    dropout & 0.0  \\
    nheads & 8  \\
    warm\_up & 1000\\
    batch\_size & 4$\times$4 \\
    \specialrule{0pt}{1pt}{1pt}
    \bottomrule
    \end{tabular}}
    \resizebox{0.491\linewidth}{!}{%
    \begin{tabular}{lc}
    \toprule
    Item  & Value \\
    \midrule
    mask\_loss & 1 \\
    obj\_loss & 1 \\
    class\_loss & 1 \\
    bbox\_loss & 5 \\ 
    giou\_loss & 2 \\
    obj\_cost & 2 \\
    class\_cost & 2 \\
    class\_cost & 2 \\
    bbox\_cost & 5 \\
    giou\_cost & 2 \\
    inner\_cost & 9999 \\
    focal\_alpha & 0.25 \\
    transformer\_activation & relu \\
    num\_queries & 400 \\
    \specialrule{0pt}{1pt}{1pt}
    \bottomrule
    \end{tabular}}
    \end{spacing}
    \caption{All configurations of SAP-DETR}
    \label{tab:configuration}%
\end{table}%

\vspace{0.1cm}

\begin{figure*}[t]
    \centering
    \vspace{-0.2cm}
    \subfloat[Comparison on COCO validation image \#159311]{
    \includegraphics[width=0.788\linewidth]{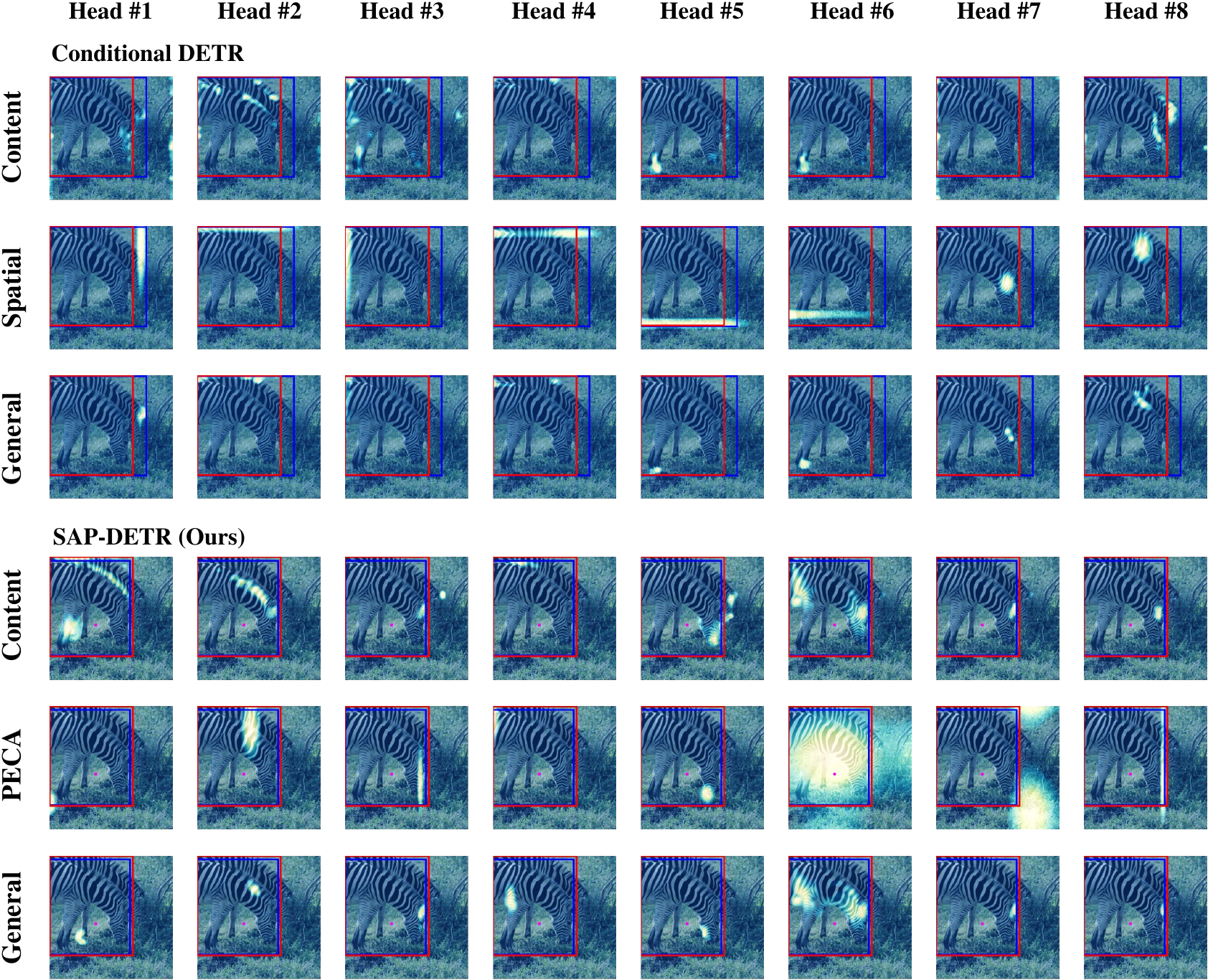}
    }\\ \vspace{0.4cm}
    \subfloat[Comparison on COCO validation image \#507975]{
    \includegraphics[width=0.788\linewidth]{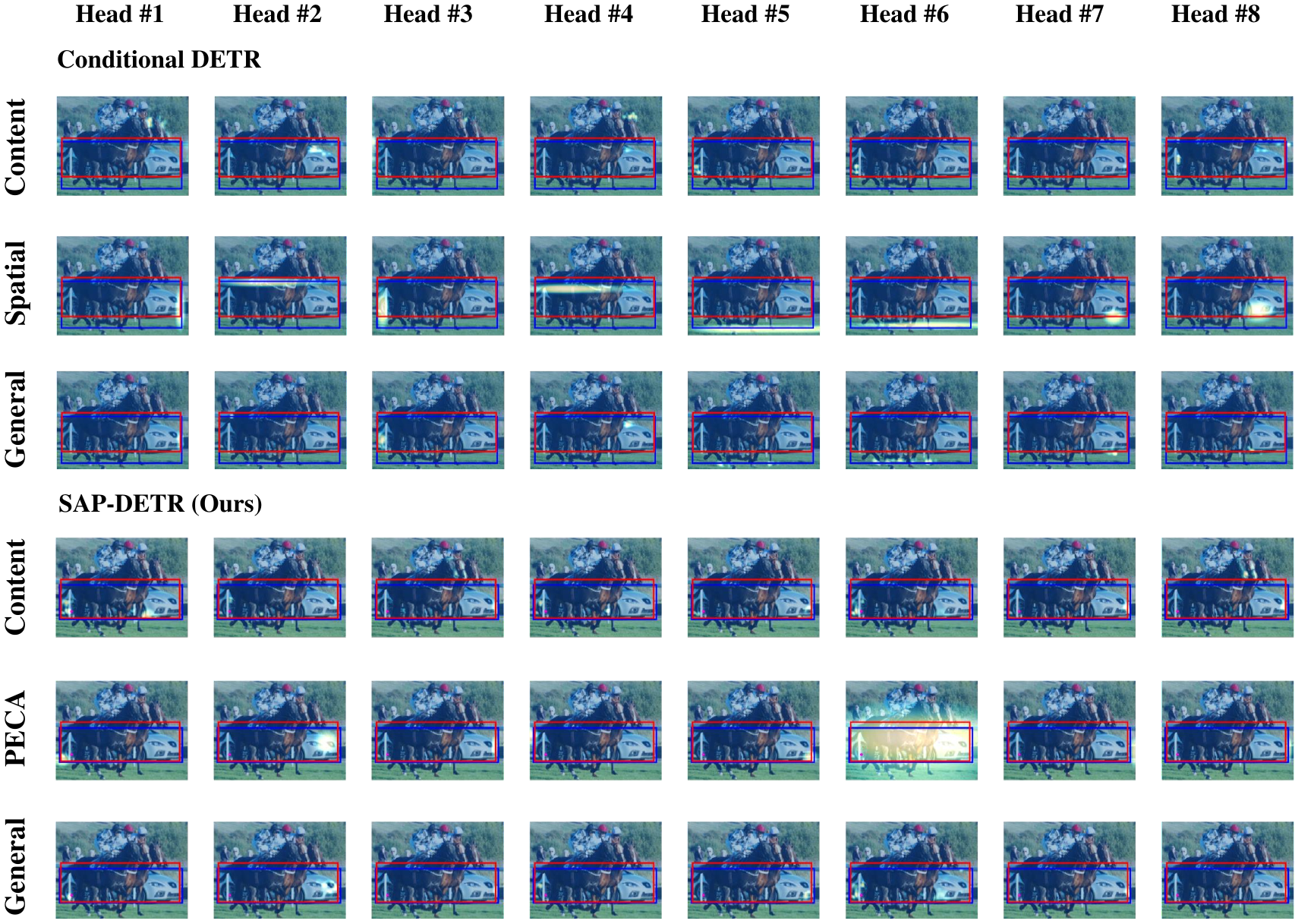}
    }
    \caption{Comparison of PECA between Conditional DETR and SAP-DETR.}
    \label{fig:sap_attn_map}
\end{figure*}

\noindent \textbf{Visualization of PECA.}
\cref{fig:sap_attn_map} visualizes both content and side attention generated by the proposed PECA. For each positive object query, we visualize each head attention map from the cross-attention mechanism. Then we compare them with the conditional spatial cross-attention. All models are based on ResNet-50 and 6-layer encoder-decoder structure under 50 training epochs. Intuitively, our content attention region mostly falls within the foreground content features, whereas a proportion of the head of Conditional DETR focus on the background. For the side attention, the attention maps of Conditional DETR are inaccurate, with several attention regions outside the bounding box. These inaccurate regions make it fail to locate the extremities efficiently and accurately. The visualization proves the effectiveness of PECA for extreme region attention and partial object detection. 

\end{document}